\documentclass[conference]{IEEEtran}
\IEEEoverridecommandlockouts
\usepackage{amsmath,amssymb,amsfonts,stmaryrd}
\usepackage{algorithmic}
\usepackage{graphicx}
\usepackage{textcomp}
\usepackage{xcolor}
\usepackage[numbers]{natbib}
\usepackage{algorithm}

\def\BibTeX{{\rm B\kern-.05em{\sc i\kern-.025em b}\kern-.08em
    T\kern-.1667em\lower.7ex\hbox{E}\kern-.125emX}}
\begin{document}

\title{Cost-Sensitive Deep Learning\\ with Layer-Wise Cost Estimation}

\author{\IEEEauthorblockN{Yu-An Chung}
\IEEEauthorblockA{
\textit{Massachusetts Institute of Technology}\\
Cambridge, MA, U.S.A.\\
andyyuan@mit.edu
}
\and
\IEEEauthorblockN{Shao-Wen Yang}
\IEEEauthorblockA{
\textit{Amazon}\\
Seattle, WA, U.S.A. \\
swyang@amazon.com
}
\and
\IEEEauthorblockN{Hsuan-Tien Lin}
\IEEEauthorblockA{
\textit{National Taiwan University}\\
Taipei, Taiwan \\
htlin@csie.ntu.edu.tw}
}

\maketitle

\begin{abstract}
While deep neural networks have succeeded in several applications, such as image classification, object detection, and speech recognition, by reaching very high classification accuracies, it is important to note that many real-world applications demand varying costs for different types of misclassification errors, thus requiring cost-sensitive classification algorithms.
Current models of deep neural networks for cost-sensitive classification are restricted to some specific network structures and limited depth. 
In this paper, we propose a novel framework that can be applied to deep neural networks with any structure to facilitate their learning of meaningful representations for cost-sensitive classification problems.
Furthermore, the framework allows end-to-end training of deeper networks directly.
The framework is designed by augmenting auxiliary neurons to the output of each hidden layer for layer-wise cost estimation, and including the total estimation loss within the optimization objective.
Experimental results on public benchmark data sets with two cost information settings demonstrate that the proposed framework outperforms state-of-the-art cost-sensitive deep learning models.
\end{abstract}


\section{Introduction}
  Deep learning has shown great success on a broad range of applications such as image classification~\citep{krizhevsky2012imagenet,ciregan2012multi,simonyan2014very,szegedy2015going,he2016deep} and speech recognition~\citep{hinton2012deep,dahl2012context}.
  Problems in such applications belong to a large class of regular classification in which the main evaluation metric is accuracy, implying each type of misclassification error is penalized equally.

  Nevertheless, using accuracy as the evaluation metric for learning does not always produce the most useful classification system in the real world.
  In fact, many real-world applications~\citep{tan1993cost,chan1998toward,fan2000multiple,zhang2010cost,TJ2011} demand varying costs for different types of misclassification errors.
  For example, different costs are useful for building a realistic face recognition system~\citep{zhang2010cost,lu2010cost,zhang2014cost,zhang2016cost}, in which a government staff being misrecognized as an impostor causes only a slight inconvenience; however, an imposer misrecognized as a staff can result in serious damage.
  Even in a simple digit recognition task, varying costs can be helpful in representing the nature of the task, as it is common and understandable to classify an ill-written~$7$ as~$1$ but classifying a~$7$ as a~$4$ would be laughable.
  Such real-world applications call for cost-sensitive classification algorithms, which aim to identify the best classifier under the application-demanded costs.

  Much research effort has been made to study cost-sensitive classification algorithms.
  In the works of~\cite{kukar1998cost,domingos1999metacost,zadrozny2001learning}, the researchers proposed to equip probabilistic classifiers with Bayes decision theory to enable the classifiers to consider the cost information during prediction.
  Some other studies extended existing cost-insensitive classification algorithms to be cost-sensitive, such as support vector machine~\citep{HT2010} or neural network~\citep{kukar1998cost,zhou2006training}.
  Recently, as deep neural networks~(DNN) have become state-of-the-art on a broad range of machine learning applications~\citep{dahl2012context,hinton2012deep,simonyan2014very,szegedy2015going,he2016deep}, researchers are attempting to make DNN cost-sensitive~\citep{YC2016}.

  One successful DNN for cost-sensitive classification, called Cost-Sensitive DNN~(CSDNN), has been recently proposed by~\cite{YC2016}.
  The training process of CSDNN consists of two steps.
  The first step is to initialize the DNN by layer-wise pretraining using a cost-sensitive variant of the conventional auto-encoder~\citep{bengio2009learning}.
  The second step involves the fine-tuning of the DNN with a loss function that incoporates the cost information.
  The final CSDNN is thus cost-sensitive in both pretraining and training stages, and is shown to be a state-of-the-art algorithm that outperforms other existing cost-sensitive classification algorithms and some deep learning alternatives.
 
  While CSDNN is state-of-the-art, its design is based on the conventional fully-connected DNN with sigmoid activation functions and experiences two issues.
  First, the design restricts the applicability to more modern structures such as convolutional~\citep{lecun1998gradient,krizhevsky2012imagenet} and pooling layers.
  Second, the sigmoid activation function suffers from the problem of diminishing gradients when the network deepens, even after careful pretraining.

  In this paper, we resolve these issues by proposing a novel framework for cost-sensitive deep learning.
  To build a cost-sensitive DNN for a~$K$-class cost-sensitive classification problem, the proposed framework replaces the layer-wise pretraining step with layer-wise cost estimation, in which~$K$ additional neurons are added to the output of each hidden layer.
  These~$K$ additional neurons serve as auxiliary units that help the DNN learn meaningful representations towards estimating the costs in each layer.
  The DNN is then trained by solving a joint optimization problem on the weighted sum of the loss functions associated with the auxiliary units.
  Experiments conducted on four benchmark data sets and two cost information settings validate that the proposed framework outperforms CSDNN.
  Furthermore, we show that the proposed framework can be easily and effectively attached to deep neural networks with ReLU~\citep{nair2010rectified} activation functions or convolutional neural networks like AlexNet~\citep{krizhevsky2012imagenet}, demonstrating that the framework provides a more general end-to-end solution for cost-sensitive deep learning than CSDNN.
  The benefits of performance and generality make the proposed framework a favorable choice in practice.

  The idea of using additional neurons as auxiliary units has been studied by several existing deep learning works, such as the well-known GoogLeNet~\citep{szegedy2015going}, which takes the additional neurons as intermediate classifiers in selected hidden layers as regularizers.
  Deeply-Supervised Nets~\citep{lee2015deeply} adds additional neurons as intermediate classifiers to all hidden layers and reported that the nodes not only serve as regularizers but also allow improved convergence behavior.
  BranchyNet~\citep{teerapittayanon2016branchynet} considers auxiliary neurons at hidden layers as early exit points of prediction to speed up testing time.
  Nevertheless, all the previous works focus on using regular (cost-insensitive) classifiers as auxiliary units.
  To the best of our knowledge, our proposed framework is the first work that tackles cost-sensitive deep learning with layer-wise auxiliary units.

  The remainder of the paper is organized as follows.
  In Section~\ref{sec:pre}, we formally define the cost-sensitive classification problem and introduce related works.
  Then, we illustrate our proposed framework in Section~\ref{sec:model}, and validate the framework with real-world data sets in Section~\ref{sec:exp}.
  Finally, we conclude in Section~\ref{sec:con}.

\section{Preliminary}
  \label{sec:pre}
  We start by formalizing the regular (cost-insensitive) classification problem and extend it to the cost-sensitive setting in Section~\ref{sec:CSC}.
  We then introduce some important deep learning research conducted to tackle cost-sensitive classification in Section~\ref{sec:dl-csc}.

\subsection{Cost-Sensitive Classification}
  \label{sec:CSC}
  In a~$K$-class regular classification problem, a size-$N$ training set $S = \{(\mathbf{x}_{n}, y_{n})\}_{n = 1} ^ {N}$ is given, where each input vector $\mathbf{x}_{n}$ is within an input space $\mathcal{X}\subseteq \mathbb{R} ^ {D}$, and each label $y_{n}$ is within a label space $\mathcal{Y} = \{1, 2, ..., K\}$.
  Regular classification aims at using~$S$ to train a classifier $g\colon \mathcal{X} \rightarrow \mathcal{Y}$ such that the expected error $\llbracket y\neq g(\mathbf{x})\rrbracket$ on the test examples~$(\mathbf{x}, y)$ is small.\footnote{The boolean operation $\llbracket \cdot \rrbracket$ is 1 if the condition is true, and 0 otherwise.}
  That is, each type of misclassification error is charged with the same penalty.

  Cost-sensitive classification extends regular classification by penalizing each type of misclassification error differently according to some given costs.
  We consider a general cost-vector setting~\citep{kukar1998cost,HT2010} when designing the proposed framework.
  The cost-vector setting represents the cost information by coupling an additional cost vector $\mathbf{c} \in [0, \infty) ^ {K}$ with each example $(\mathbf{x}, y)$, where the $k$-th component $\mathbf{c}[k]$ of the cost vector $\mathbf{c}$ denotes the cost of predicting $\mathbf{x}$ as class $k$, and naturally $\mathbf{c}[y] = 0$.
  Consider a cost-sensitive training set $S_{c} = \{(\mathbf{x}_{n}, y_{n}, \mathbf{c}_{n})\}_{n = 1} ^ {N}$, cost-sensitive classification aims at using $S_{c}$ to train a classifier $g_{c}\colon \mathcal{X} \rightarrow \mathcal{Y}$ such that the expected cost $\mathbf{c}[g_{c}(\mathbf{x})]$ on the test examples~$(\mathbf{x}, y, \mathbf{c})$ is small.

  A special case of the cost-vector setting is the cost-matrix setting, where the cost information is encoded by a $K \times K$ cost matrix~$\mathbf{C}$ and each entry $\mathbf{C}(y, k) \in [0, \infty)$ indicates the cost for predicting a class-$y$ example as class~$k$.
  The information within the cost matrix can be simply cast as the cost vectors by defining the cost vector in $(\mathbf{x}, y, \mathbf{c})$ as the $y$-th row of the cost matrix $\mathbf{C}$.
  The cost-matrix setting, albeit less general, allows real-world applications to specify their demanded costs more easily.
  We follow many earlier cost-sensitive classification works~\citep{kukar1998cost,domingos1999metacost,abe2004iterative,HT2010,YC2016} to take the cost-matrix setting when conducting benchmark experiments.

\subsection{Deep Learning for Cost-Sensitive Classification}
  \label{sec:dl-csc}
  Nowadays, most DNNs are designed to solve regular classification problems~\citep{simonyan2014very,szegedy2015going,he2016deep}.
  Those DNNs usually consist of several hidden layers with a softmax layer of~$K$ neurons at the end.
  Each input vector~$\mathbf{x}$ propagates through different hidden layers and is transformed into different levels of latent representations.
  The softmax layer converts the last latent representation into per-class probability estimation, and takes the class with the highest estimated probability as the prediction $g(\mathbf{x})$ of the network.

  On the other hand, only few works have explored cost-sensitive classification using shallow or deep neural networks.
  Prior to the prevailing of deep neural networks, \cite{zhou2006training} pioneered the study of making neural networks cost-sensitive by sampling and threshold-moving to tackle the class imbalance problem;
  \cite{kukar1998cost} proposed four approaches of modifying neural networks for cost-sensitivity.
  Recently, \cite{YC2016} attempted to make cost-sensitive neural networks deeper, and proposed a cost-sensitive deep learning algorithm called Cost-Sensitive DNN~(CSDNN).
  In terms of the network structure, CSDNN starts with a regular DNN with fully-connected layers, but replaces the softmax layer at the end of the DNN by a cost-estimation layer.
  Each of the~$K$ neurons in the cost-estimation layer provides per-class cost estimation with regression instead of per-class probability estimation.
  Then, the class with the lowest estimated cost can be naturally taken as the prediction $g_c(\mathbf{x})$ of the network.
  \cite{YC2016} proposed to train the structure with a cost-sensitive loss function $L_{\mathrm{OSR}}$ on the cost-estimation layer.%
  \footnote{The term $L_{\mathrm{OSR}}$ stands for One-Sided Regression and roots from a cost-sensitive SVM work~\citep{HT2010}. Details are omitted here for lack of space.}

  \cite{YC2016} then found that the performance of the network can be further improved by careful pretraining, and proposed a Cost-Sensitive Auto-Encoder~(CSAE) to pretrain the structure above in a layer-wise manner.
  CSAE operates similar to a conventional auto-encoder~\citep{bengio2009learning}, which is a shallow neural network that maps any input $\mathbf{x}$ to a representation such that the output $\tilde{\mathbf{x}}$ is a close reconstruction of the original input.
  The reconstruction error is commonly measured by cross-entropy loss, denoted by $L_{\mathrm{CE}}$.
  What makes CSAE different is that the shallow network is augmented with~$K$ additional output neurons for cost estimation.
  That is, CSAE attempts to not only reconstruct $\mathbf{x}$ but also digest the cost information by estimating the cost vector $\mathbf{c}$.
  The attempt is represented with a mixture loss $(1\!-\!\beta) \cdot L_{\mathrm{CE}} + \beta \cdot L_{\mathrm{OSR}}$ with a balancing coefficient $\beta \in [0, 1]$ on the output layer of CSAE.
  When $\beta\!=\!0$, CSAE degrades to a conventional auto-encoder.

\begin{figure}
    \centering
    \includegraphics[scale=0.25]{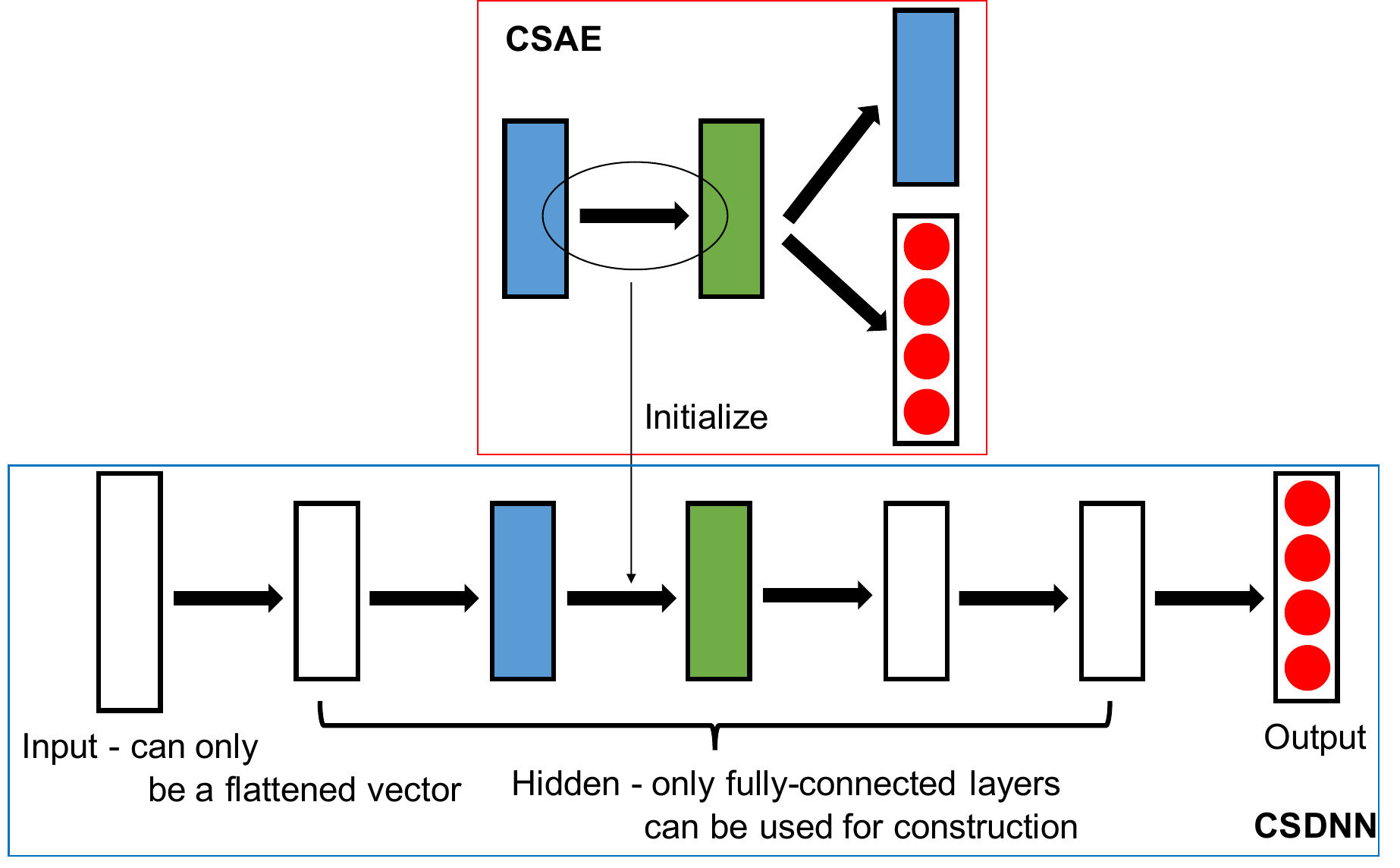}
    \caption{
      CSAE pretraining for CSDNN~\citep{YC2016}
    }
    \label{fig:CSAE_CSDNN}
  \end{figure}

  Figure~\ref{fig:CSAE_CSDNN} illustrates how CSAE is used to pretrain CSDNN.
  With the pretraining, each layer of (initial) latent representations in CSDNN carries some ability to estimate the costs.
  That is, the pretraining makes the latent representations \textit{cost-aware}.
  \cite{YC2016} reported that such initialization indeed allows CSDNN to converge to a better optima and to reach state-of-the-art performance.

\section{Proposed Framework}
\label{sec:model}
  While CSDNN is state-of-the-art, its design is based on fully-connected layers with sigmoid activation functions and thus suffers from some issues. Next, we discuss the issues behind CSDNN that motivate us 
  to study a general framework that allows conducting cost-sensitive deep learning more effectively and effortlessly in Section~\ref{sec:drawbacks}.
  Then we present our proposed framework in Section~\ref{sec:csaux}.

\subsection{Motivation}
  \label{sec:drawbacks}
  Arguably the key idea within CSDNN is pretraining with CSAE.
  To understand the issues behind CSDNN, we first review the necessity of pretraining for general deep learning.
  In earlier years, neural networks used sigmoid or hyperbolic-tangent activation functions for non-linear transformation in the hidden layers~\citep{dayhoff1990intro,lawrence1997face,hinton2006fast,bengio2009learning}.
  Both functions, which exhibit flatness in part of their curves, can cause the gradients of the network to be small.
  As the depth of the network increases, the small gradients in the latter layers of the network make the gradients in the earlier layers even smaller during back-propagation, a phenomenon known as the diminishing gradients.
  The phenomenon results in poor local optima of the entire network, and restricts the depth of earlier neural networks~\citep{michael2015nndl}.
  Earlier works by \cite{hinton2006fast} and \cite{bengio2009learning} tackled the diminishing-gradient problem by proposing a greedy layer-wise pretraining strategy with Restricted Boltzmann Machines and auto-encoders for initializing the weights of DNN.
  Pretraining helped mitigate the problem to some degree, but the problem would resurface as the network deepens if we stick with the same activation functions.

  In recent years, another route to resolve the diminishing-gradient problem is to consider other activation functions, such as the rectifier linear unit~(ReLU)~\citep{nair2010rectified}.
  As ReLU does not suffer from the diminishing-gradient problem as much as sigmoid or hyperbolic-tangent activation functions, pretraining is no longer necessary~\citep{glorot2011deep}.
  Nowadays, ReLU and many of its variants~\citep{xu2015empirical,he2015delving} become the mainstream activation functions in modern deep learning studies~\citep{maas2013rectifier,he2016deep}.

  CSDNN~\citep{YC2016} intended to conduct cost-sensitive deep learning by mimicking what~\cite{bengio2009learning} did for regular deep learning: using sigmoid activation functions, and adopting greedy laywer-wise pretraining.
  Thus, CSDNN carries the same problem of diminishing gradients when the network deepens, as our experimental results in Section~\ref{sec:exp} will demonstrate.
  To keep cost-sensitive deep learning up to date with modern deep learning studies, it is then necessary to conduct cost-sensitive deep learning with other routes, such as adopting ReLU and removing the pretraining stage.

  Nevertheless, directly removing the pretraining stage in CSDNN throws away one important benefit of CSAE in making the latent representations cost-aware.
  Next, we present our proposed framework to rescue the benefit.
  As we shall demonstrate later, the proposed framework carries an additional advantage of being generally applicable to a wide range of network structures and activation functions, including but not limited to ReLU.

\subsection{Layer-Wise Cost Estimation}
  \label{sec:csaux}
  \begin{figure}
    \centering
    \includegraphics[scale=0.25]{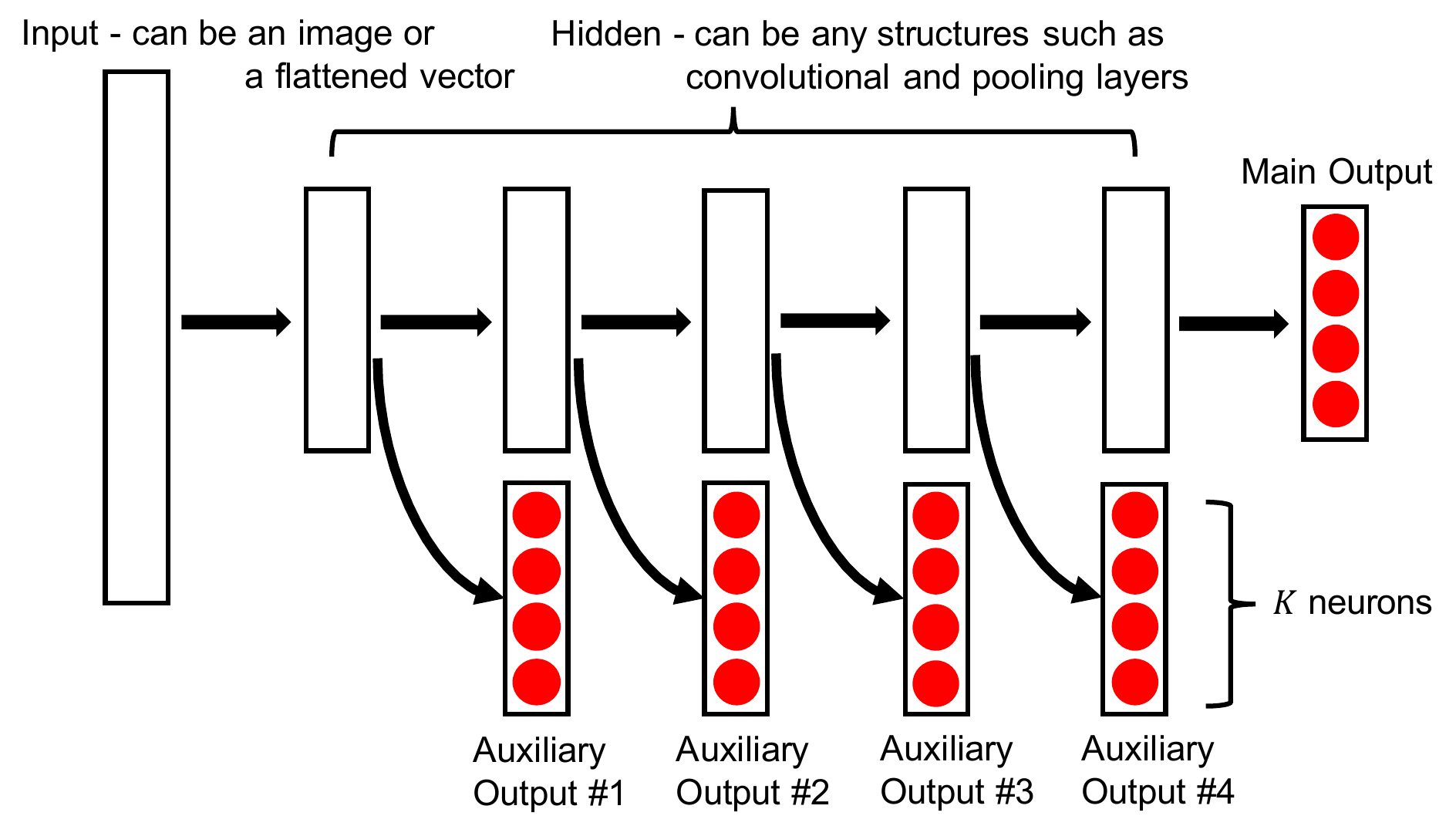}
    \caption{
      a DNN with five hidden layers dressed with the proposed Auxiliary Cost-Sensitive Targets~(AuxCST) framework
    }
    \label{fig:AuxCST_example}
  \end{figure}

  Our key goal is to construct a DNN that can simultaneously enjoy the benefit of cost-aware representation extraction (similar to that provided by CSAE), and the flexibility of using any structures.
  CSAE achieved cost-aware representation extraction by using $K$ additional neurons in the auto-encoder for cost estimation.
  Our key idea is to also use $K$ additional neurons for cost estimation, but instead of adding them to the auto-encoders that are separated from the DNN, we propose to directly put $K$ neurons into each layer of the DNN.
  That is, we propose to replace CSAEs by ``merging'' their additional neurons with the DNN of our interest.
  The proposed structure is illustrated with Figure~\ref{fig:AuxCST_example}.
  By dressing the original DNN with $K$ additional neurons in each layer that serve as auxiliary outputs, the extracted latent representations carry some ability to estimate the costs, thus achieving cost-aware representation extraction almost effortlessly.

  As shown in Figure~\ref{fig:AuxCST_example}, in addition to augmenting $K$ additional neurons to each layer of the DNN, we follow CSDNN and replace the output layer of the DNN with a cost-estimation layer.
  Then, the only remaining task is to train the ``upgraded'' DNN with a proper loss function. We consider a simple weighted-mixture loss function of the main one-sided regression loss function at the output layer, and the auxiliary one-sided regression loss functions at the hidden layers.
  In particular, let $L^{(i)}_{\mathrm{OSR}}$ denote the auxiliary loss function for the output of the $i$-th hidden layer and $L^{(*)}_{\mathrm{OSR}}$ denote the main loss function at the output layer, we train the upgraded DNN with the loss
  \begin{equation}
    \label{eq:different_alpha}
    \sum^{H - 1}_{i = 1}\alpha_{i} \cdot L^{(i)}_{\mathrm{OSR}} + L^{(*)}_{\mathrm{OSR}},
  \end{equation}
  where $H$ is the number of hidden layers in the DNN, and $\alpha_{i}$ is the balancing coefficient for $L^{(i)}_{\mathrm{OSR}}$.%
  \footnote{There is
    no need to consider $L^{(H)}_{\mathrm{OSR}}$ for the outputs of the last hidden layer, as the main loss function $L^{(*)}_{\mathrm{OSR}}$ readily conducts cost estimation.} 

  With the proposed structural addons and the mixture loss function, we are now ready to present the full framework in Algorithm~\ref{alg:AuxCST}. The framework will be named as Auxiliary Cost-Sensitive Targets (AuxCST).
  While the novel framework appears simple, it carries many practical benefits. 
  With the framework, we can now flexibly use ReLU or other activation functions and thus avoid diminishing-gradient problem.
  We can also build cost-sensitive DNN with any structures, such as image inputs with convolutional and pooling layers.
  Furthermore, we can apply this framework directly and effortlessly on 
  any state-of-the-art DNN structures such as VGG~\citep{simonyan2014very} and ResNet~\citep{he2016deep} for solving large-scale cost-sensitive classification problems.

  \begin{algorithm}[h]
    \caption{Auxiliary Cost-Sensitive Targets (AuxCST)}
    \label{alg:AuxCST}
    \begin{algorithmic}[1]
      \renewcommand{\algorithmicrequire}{\textbf{Input:}}
      \REQUIRE your favorite regular DNN or any off-the-shelf one~\citep{krizhevsky2012imagenet,simonyan2014very,szegedy2015going,he2016deep} with $H$ hidden layers; balancing coefficients $\{\alpha_{i}\} ^ {H - 1}_{i = 1}$
      \STATE Replace the softmax layer at the end of DNN with $K$ regression neurons and loss function $L^{(*)}_{\mathrm{OSR}}$
      \FOR {$i = 1, 2, \ldots, H - 1$}
      \STATE Add $K$ additional regression neurons with loss function $L^{(i)}_{\mathrm{OSR}}$ to the output of the $i$-th hidden layer and connect them fully to the $i$-th hidden layer
      \ENDFOR
      \STATE Train the new DNN by back-propagation on~(\ref{eq:different_alpha})
    \end{algorithmic}
  \end{algorithm}

\section{Experiments}
  \label{sec:exp}
  Three sets of experiments are conducted to validate the usefulness of the proposed AuxCST framework.
  Four benchmark data sets are used for the experiments: MNIST, CIFAR-10~\citep{Krizhevsky09}, CIFAR-100~\citep{Krizhevsky09}, and Caltech-256~\citep{griffin2007caltech}.
  MNIST belongs to handwritten digit recognition task where each example is a $28 \times 28$ gray-scale digit image; CIFAR-10 is a well-known image recognition data set with 10 classes where the size of each image is $3 \times 32 \times 32$ (3 for RGB); CIFAR-100 is just like CIFAR-10, except it has 100 classes; Caltech-256 is another popular object recognition data set that contains 256 classes, and the size of each RGB image is roughly $300 \times 400$ in average.
  For all data sets, the training and testing splits follow the source websites; the input vectors in training set are linearly scaled to $[0, 1]$, and the input vectors in the testing sets are scaled accordingly.
  We use MNIST, CIFAR-10, and CIFAR-100 in our first two experiments, and Caltech-256 in our last experiment.

  The four data sets were originally collected for regular (cost-insensitive) classification and thus contain no cost information.
  We adopt the most frequently-used benchmark in cost-sensitive learning, the randomized proportional setup~\citep{abe2004iterative}, to generate the costs.
  For a regular data set $S = \{(\mathbf{x}_{n}, y_{n})\}_{n = 1} ^ {N}$, the setup first generates a $K \times K$ matrix $\mathbf{C}$, and sets the diagonal entries $\mathbf{C}(y, y)$ to 0 while sampling the non-diagonal entries $\mathbf{C}(y, k)$ uniformly from $[0, 10\frac{|\{n: y_{n} = k\}|}{|\{n: y_{n} = y\}|}]$.
  Then, for each example $(\mathbf{x}_{n}, y_{n})$ in $S$, its cost vector $\mathbf{c}_{n}$ is defined as the $y_{n}$-th row of matrix $\mathbf{C}$.
  The randomized proportional setup generates the cost information that takes the class distribution of the data set into account, charging a higher cost (in expectation) for misclassifying a minority class, and can thus be used to deal with imbalanced classification problems.

  Arguably one of the most important use of cost-sensitive classification is to deal with imbalanced data sets.
  Nevertheless, the first three data sets MNIST, CIFAR-10, and CIFAR-100 are somewhat balanced, and the randomized proportional setup may generate similar cost for each type of misclassification error.
  To better meet the real-world usage scenario and increase the diversity of data sets, we further conduct experiments to evaluate the algorithms with imbalanced data sets.
  In particular, for each of the first three data sets MNIST, CIFAR-10, and CIFAR-100, we construct a variant data set by randomly picking 40\% of the classes and removing 70\% of the examples that belong to those 40\% classes.
  We will name these imbalanced variants as $\mathrm{MNIST}_{\mathrm{imb}}$, $\mathrm{CIFAR}$-$\mathrm{10}_{\mathrm{imb}}$, and $\mathrm{CIFAR}$-$\mathrm{100}_{\mathrm{imb}}$, respectively.

  Our first experiment in Section~\ref{sec:alpha_perf_relation} intends to investigate the relationship between the balancing coefficient $\alpha_{i}$ in (\ref{eq:different_alpha}) for using AuxCST and the performance.
  Our second experiment in Section~\ref{sec:FC_compare} compares DNN equipped with AuxCST framework with state-of-the-art CSDNN~\citep{YC2016}) to show the usefulness of AuxCST.
  For the first and the second experiments, the cost information was generated by the randomized proportional setup.
  To study the effectiveness of AuxCST for other kinds of cost information, we consider a setup that generates the costs
  from the top-down hierarchical tree structure for the 256 classes of Caltech-256 data set in Section~\ref{sec:closeness}. In particular, the costs are calculated
  based on the closeness between each class within the tree and the resulting cost-sensitive classifier shall be more human mimicking by
  taking the categorical similarity into account.
  We will apply AuxCST on top of the well-known AlexNet~\citep{krizhevsky2012imagenet} to tackle this realistic cost-sensitive classification problem.
  In each of the three experiments, we will describe the goal of the experiment, present the experimental results, and provide discussions and conclusions.

\subsection{How does $\alpha_{i}$ affect AuxCST?}
  \label{sec:alpha_perf_relation}
  In our proposed Auxiliary Cost-Sensitive Targets~(AuxCST) framework, $K$ additional neurons are added in parallel to each of the hidden layer in DNN.
  As an example $\mathbf{x}$ propagates through the network, in addition to the final prediction layer, the DNN also outputs $K$ values in each hidden layer.
  Same with the final prediction layer, these additional $K$ neurons in each hidden layer also aim to estimate the per-class costs, and are coupled with $L_{\mathrm{OSR}}$.
  The final objective function for optimizing the entire DNN turns out to be a weighted sum of the main one-sided loss for the final prediction layer and the auxiliary one-sided loss for all hidden layers, and has the form~(\ref{eq:different_alpha}).

  In this experiment, we would like to investigate the relationship between the selection of $\alpha_{i}$ in~(\ref{eq:different_alpha}) and the performance (average test costs) of AuxCST framework.
  To simplify the experiment, we keep all coefficients $\alpha_{i}$ to identical values, that is, $\alpha_{1} = \alpha_{2} = ... = \alpha_{H - 1} = \alpha$, and~(\ref{eq:different_alpha}) becomes:
  \begin{equation}
    \label{eq:same_alpha}
    \alpha \cdot \sum^{H - 1}_{i = 1} L^{(i)}_{\mathrm{OSR}} + L^{(*)}_{\mathrm{OSR}},
  \end{equation}
  and we increase the value of $\alpha$ from $0$ to $1$ by a step $0.1$.
  MNIST, CIFAR-10, CIFAR-100 and their imbalanced variants are used in this experiment, and their cost information is generated by randomized proportional setup. 
  We constructed fully-connected DNN with varying numbers of hidden layers $H = \{1, 2, 3, 4, 5\}$, where each hidden layer consists of 1024 neurons.
  Note that our proposed AuxCST framework can be applied to DNN consists of any kind of layers, but since our goal in current experiment is not to pursue the best performance but to investigate more about AuxCST, we choose to use only fully-connected layers for constructing DNN in order to reduce the amount of hyper-parameters.

\begin{figure*}[!ht]
    \hspace{-0.8cm}
    \centering
    \begin{minipage}[b]{.3\textwidth}
      \includegraphics[scale=0.29]{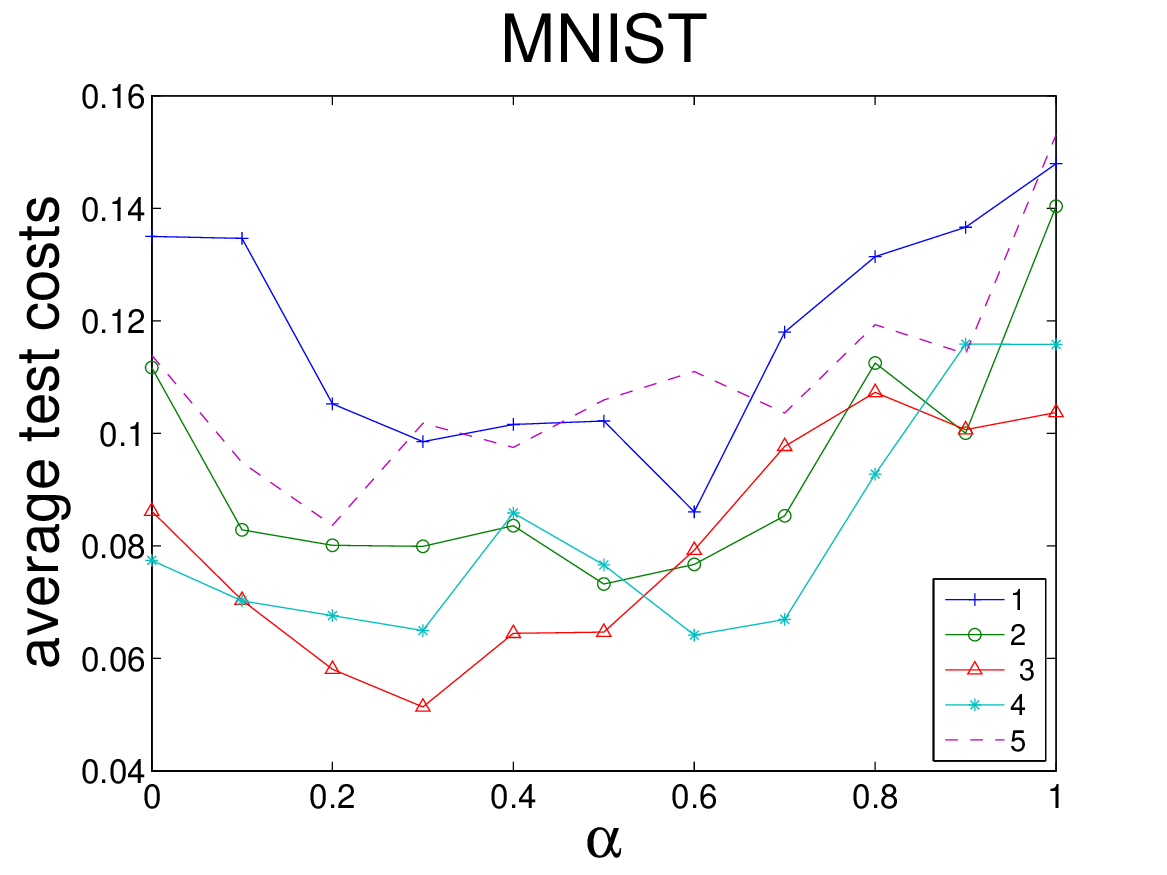}
    \end{minipage}\qquad
    \begin{minipage}[b]{.3\textwidth}
      \includegraphics[scale=0.29]{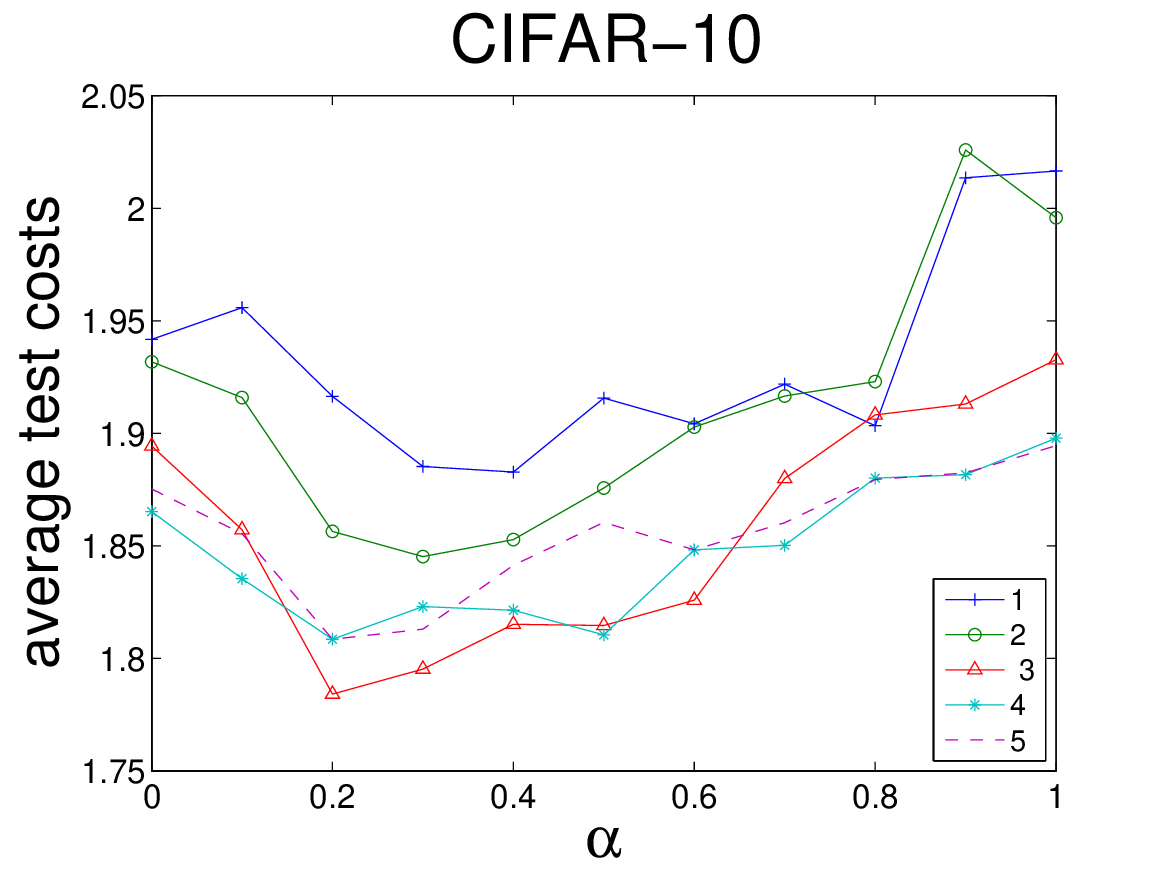}
    \end{minipage}\qquad
    \begin{minipage}[b]{.3\textwidth}
      \includegraphics[scale=0.29]{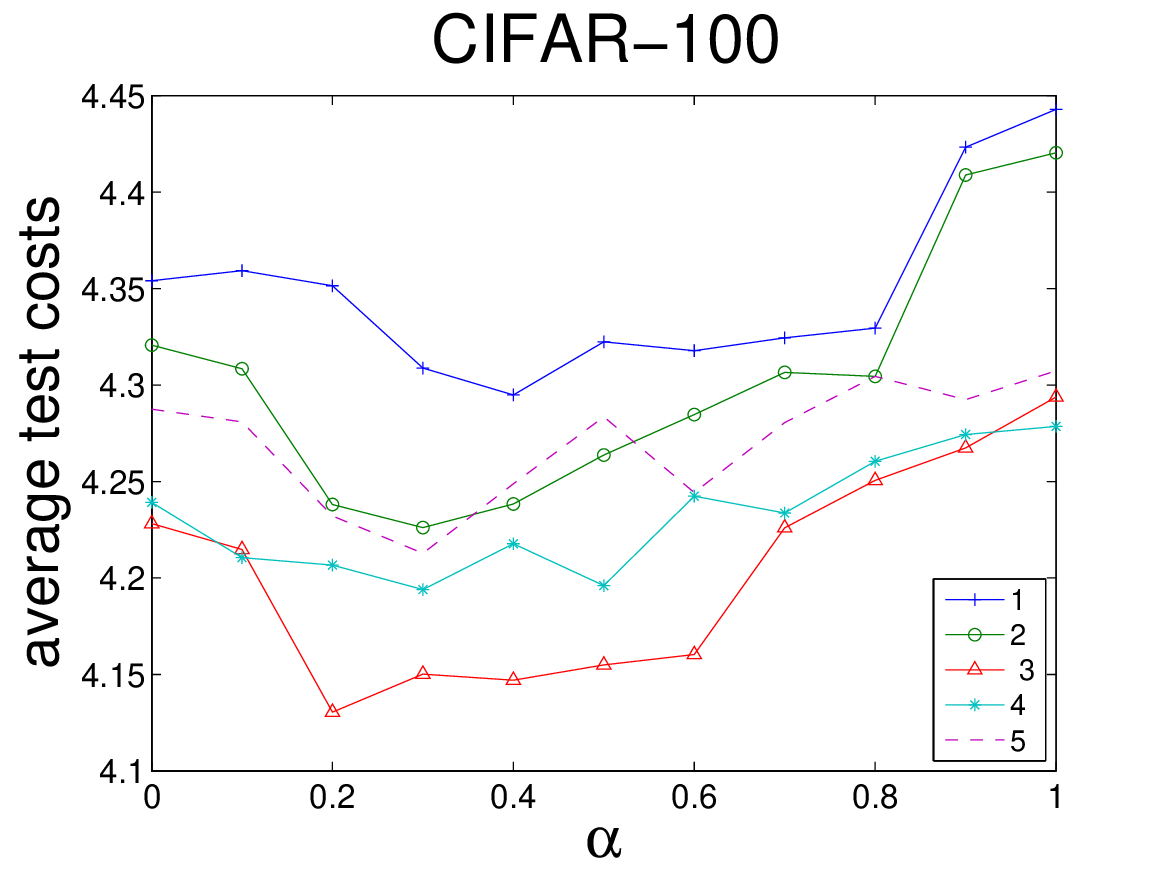}
    \end{minipage}\qquad \\
    \hspace{-0.8cm}
    \begin{minipage}[b]{.3\textwidth}
      \includegraphics[scale=0.29]{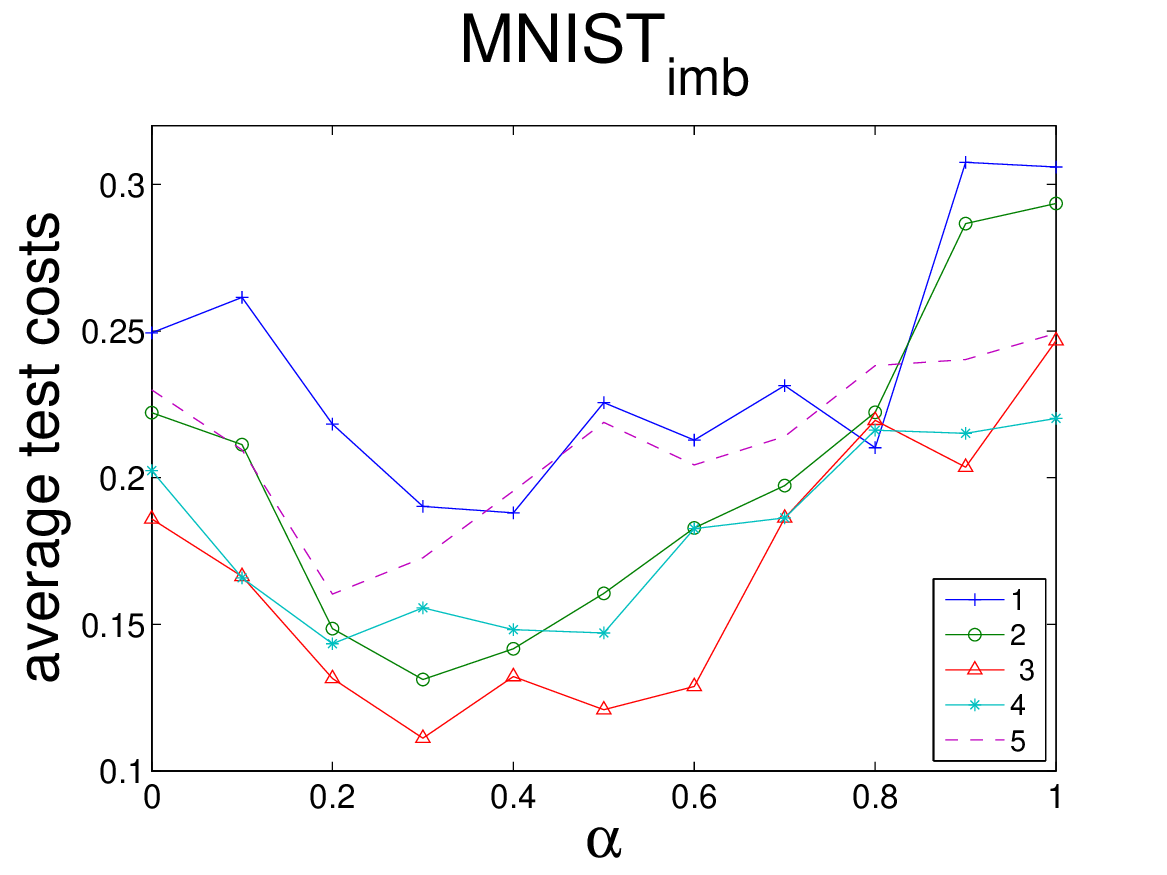}
    \end{minipage}\qquad
    \begin{minipage}[b]{.3\textwidth}
      \includegraphics[scale=0.29]{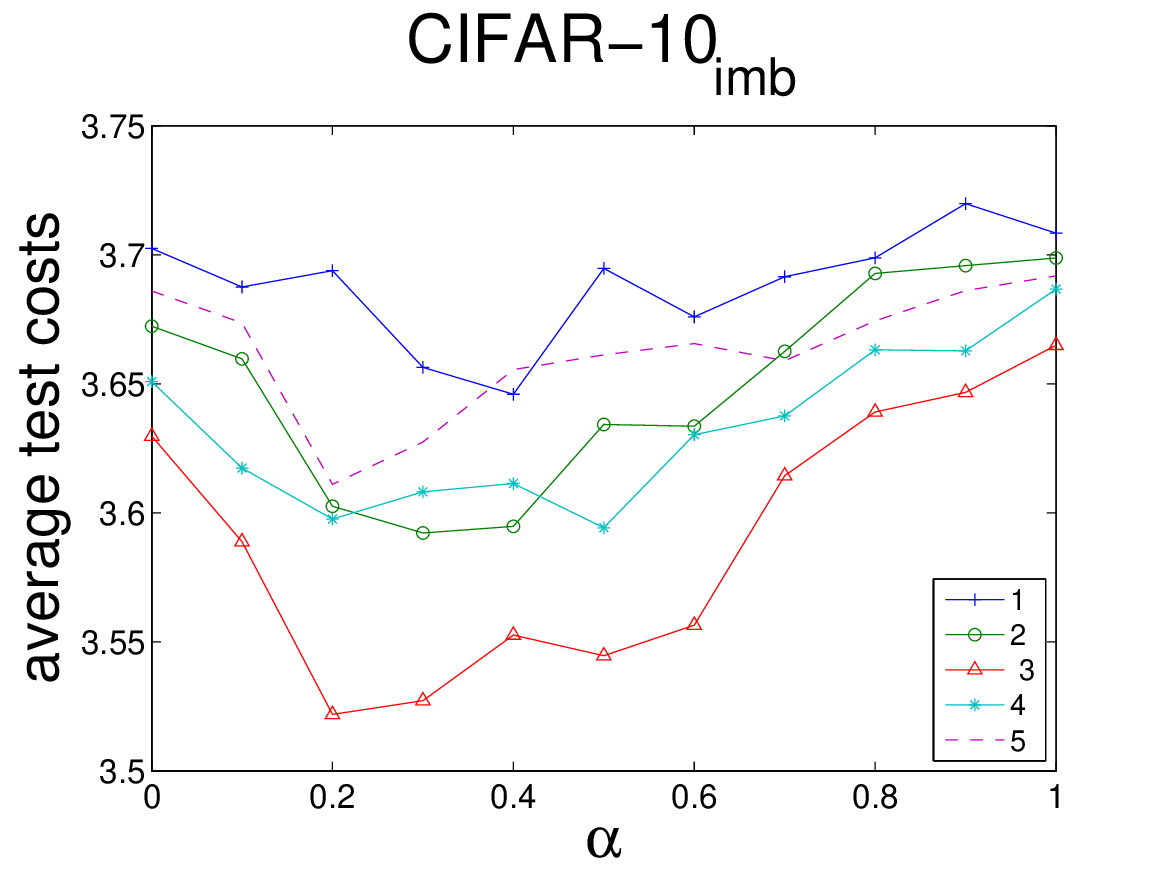}
    \end{minipage}\qquad
    \begin{minipage}[b]{.3\textwidth}
      \includegraphics[scale=0.29]{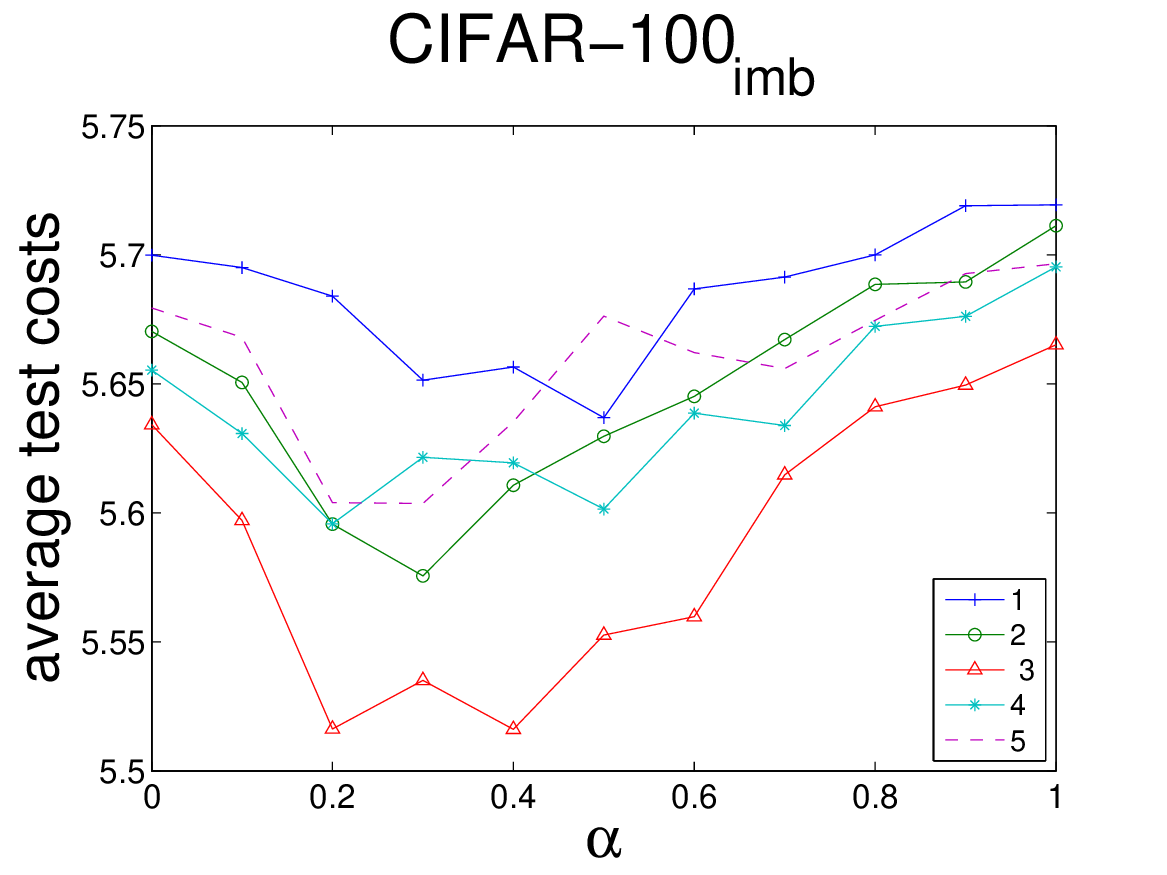}
    \end{minipage}\qquad
    \caption{
      The figure shows the results of our first experiment described in Section~\ref{sec:alpha_perf_relation}.
      The six sub-figures show the relationship between the selection of $\alpha$ in~(\ref{eq:same_alpha}) for using AuxCST framework and the performance achieved on MNIST, CIFAR-10, CIFAR-100, $\mathrm{MNIST}_{\mathrm{imb}}$, $\mathrm{CIFAR}$-$\mathrm{10}_{\mathrm{imb}}$, and $\mathrm{CIFAR}$-$\mathrm{100}_{\mathrm{imb}}$.
      Each figure contains five curves, the numbers in the legend are the number of hidden layers, and each number corresponds to a curve.
      The x-axis is the value of $\alpha$, and the y-axis is the corresponding average test costs.
    }
    \label{fig:alpha_plot}
  \end{figure*}

  The experimental results are shown in Figure~\ref{fig:alpha_plot}.
  For each data set, we plot 5 curves (because we tested with 5 kinds of numbers of hidden layers), where the x-axis is the value of $\alpha$, and the y-axis is the corresponding average test costs achieved.
  Note that when $\alpha = 0$, it means that the DNN does not make use of AuxCST framework.
  From the six figures, no matter how many hidden layers there are, roughly U-shaped curves could be observed, and the lowest average test costs were achieved when $\alpha$ fell in the range $0.2 \sim 0.5$, implying that $\alpha$ within this range best balanced~(\ref{eq:same_alpha}).
  Another phenomenon we can observe from these six figures is that, when the number of hidden layers increased from 1 to 3, the performance was improved as the entire curve moved downward; but when the number of hidden layers continued to increase from 3 to 5, the performance could not be further improved and even got worse.
  Such phenomenon could probably be attributed to overfitting - the training examples of MNIST, CIFAR-10, and CIFAR-100 are only 60K, 50K, and 50K, respectively, and their imbalanced counterparts contain even less examples due to the removals, so using more than 3 hidden layers would be overkilling.

\subsection{Compare with state-of-the-art}
  \label{sec:FC_compare}
  In this experiment, we build two DNNs with and without AuxCST framework and compare them to state-of-the-art Cost-sensitive Deep Neural Network~(CSDNN)~\citep{YC2016}.
  We emphasize the two major drawbacks of CSDNN here:
  \begin{enumerate}
    \item CSDNN uses sigmoid functions for non-linear transformations, and this will eventually results in diminishing gradients when the network grows deeper.
    \item CSDNN can be applied to DNN that consists of only fully-connected layers, this puts limits on its potential to be extended and applied to more challenging tasks that require modern neural components such as convolution and pooling layers.
  \end{enumerate}
  To give CSDNN a fair chance of comparison, the two DNNs we build also consist of only fully-connected layers, and ReLU is used as activation function.
  The first DNN is equipped with AuxCST by setting $\alpha_{i} = 0.2$, as $0.2$ was found to be one of the best value balancing for~(\ref{eq:same_alpha}) in Section~\ref{sec:alpha_perf_relation}, we will refer to this DNN as AuxDNN.
  The second DNN, which will be referred to as NaiveDNN, did not make use of AuxCST and was directly optimized by $L_{\mathrm{OSR}}$, it is equivalent to setting $\alpha = 0$ in~(\ref{eq:same_alpha}).

\begin{figure*}[!ht]
    \hspace{-0.8cm}
    \centering
    \begin{minipage}[b]{.3\textwidth}
      \includegraphics[scale=0.29]{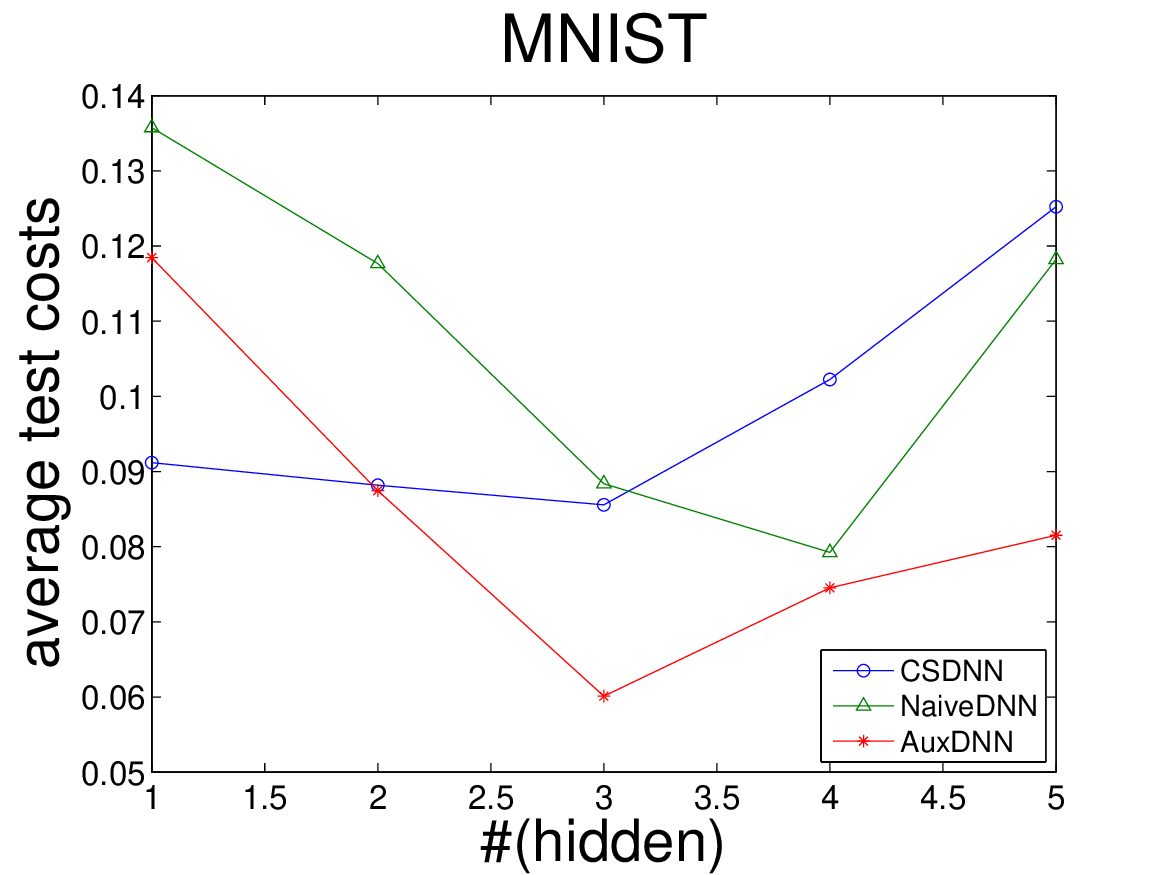}
    \end{minipage}\qquad
    \begin{minipage}[b]{.3\textwidth}
      \includegraphics[scale=0.29]{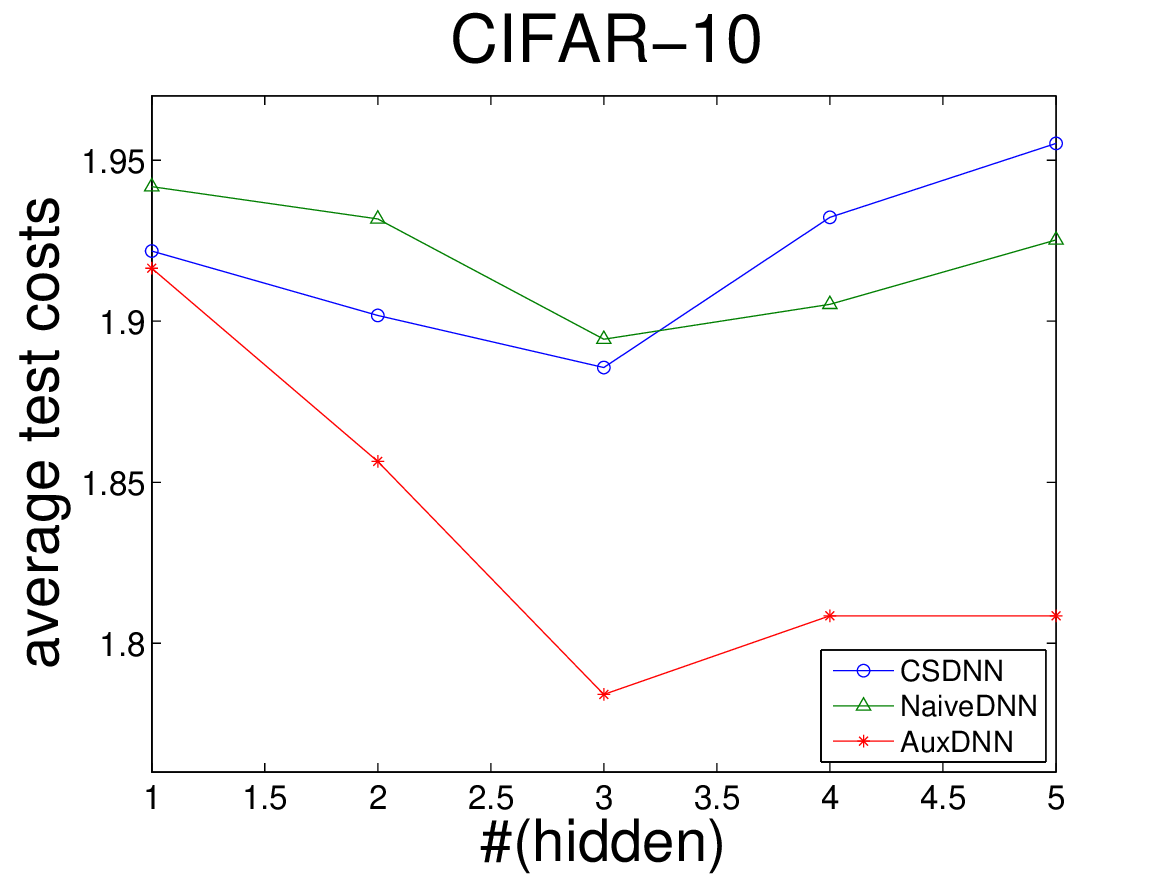}
    \end{minipage}\qquad
    \begin{minipage}[b]{.3\textwidth}
      \includegraphics[scale=0.29]{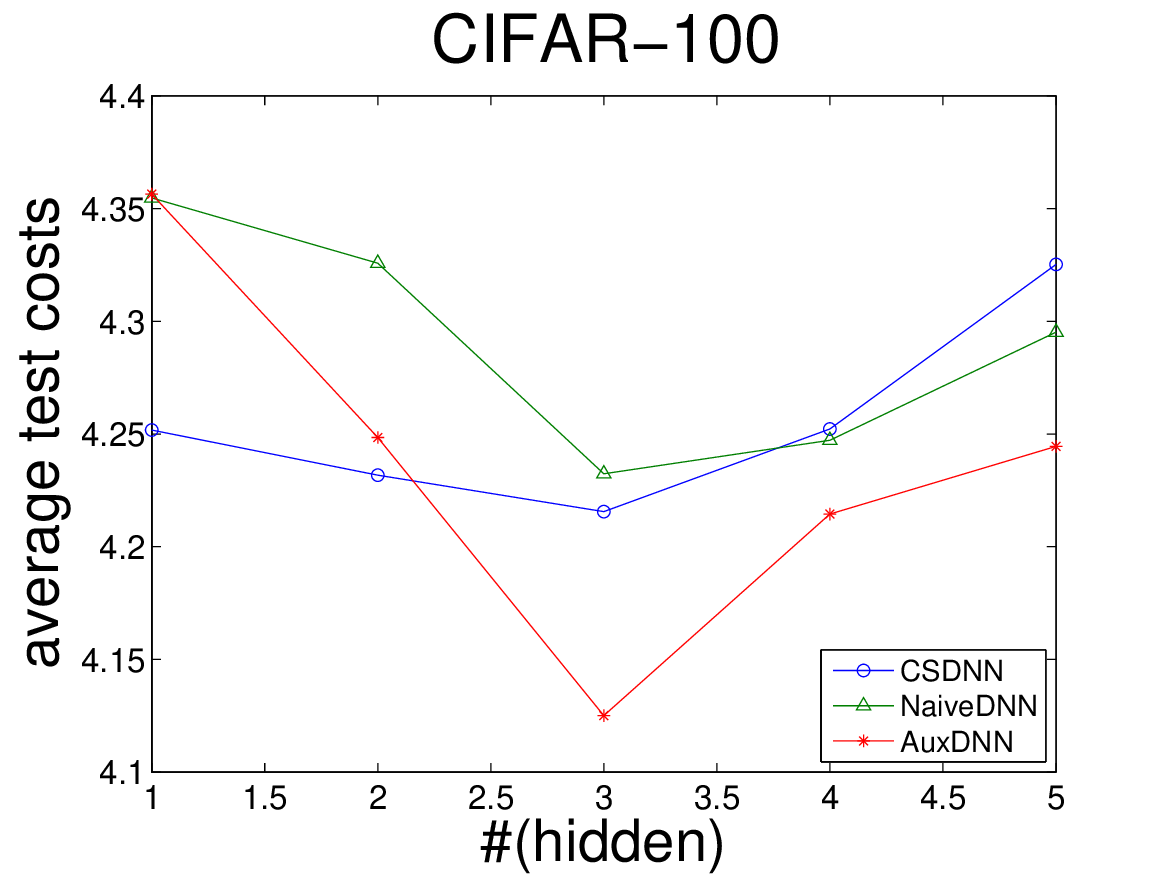}
    \end{minipage}\qquad \\
    \hspace{-0.8cm}
    \begin{minipage}[b]{.3\textwidth}
      \includegraphics[scale=0.29]{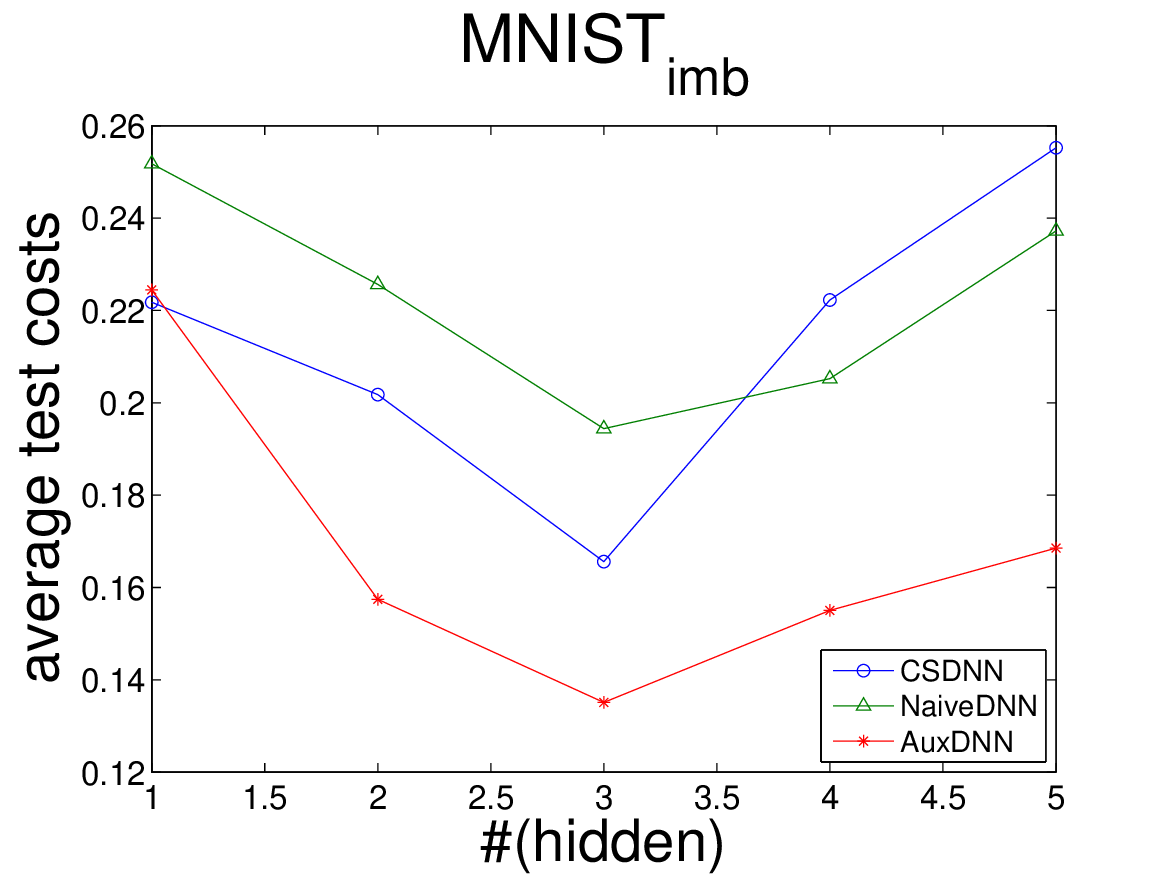}
    \end{minipage}\qquad
    \begin{minipage}[b]{.3\textwidth}
      \includegraphics[scale=0.29]{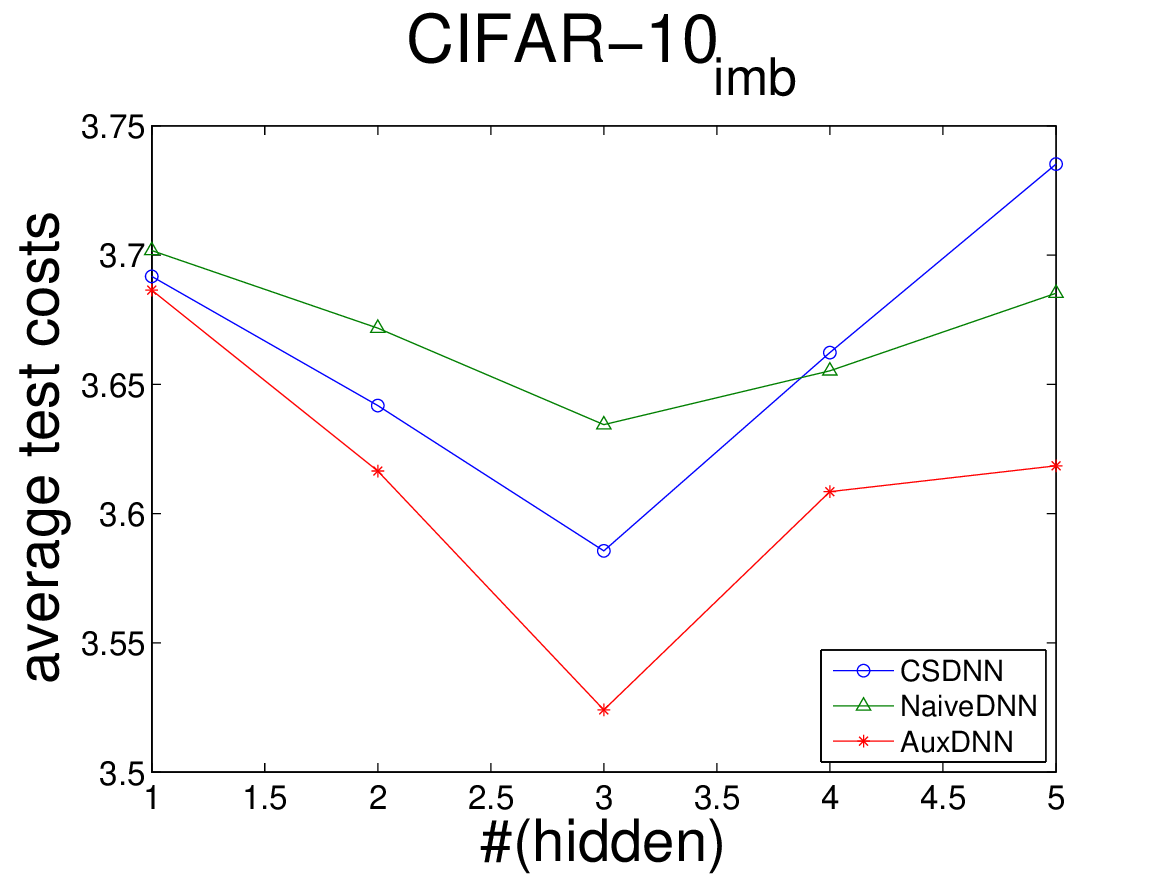}
    \end{minipage}\qquad
    \begin{minipage}[b]{.3\textwidth}
      \includegraphics[scale=0.29]{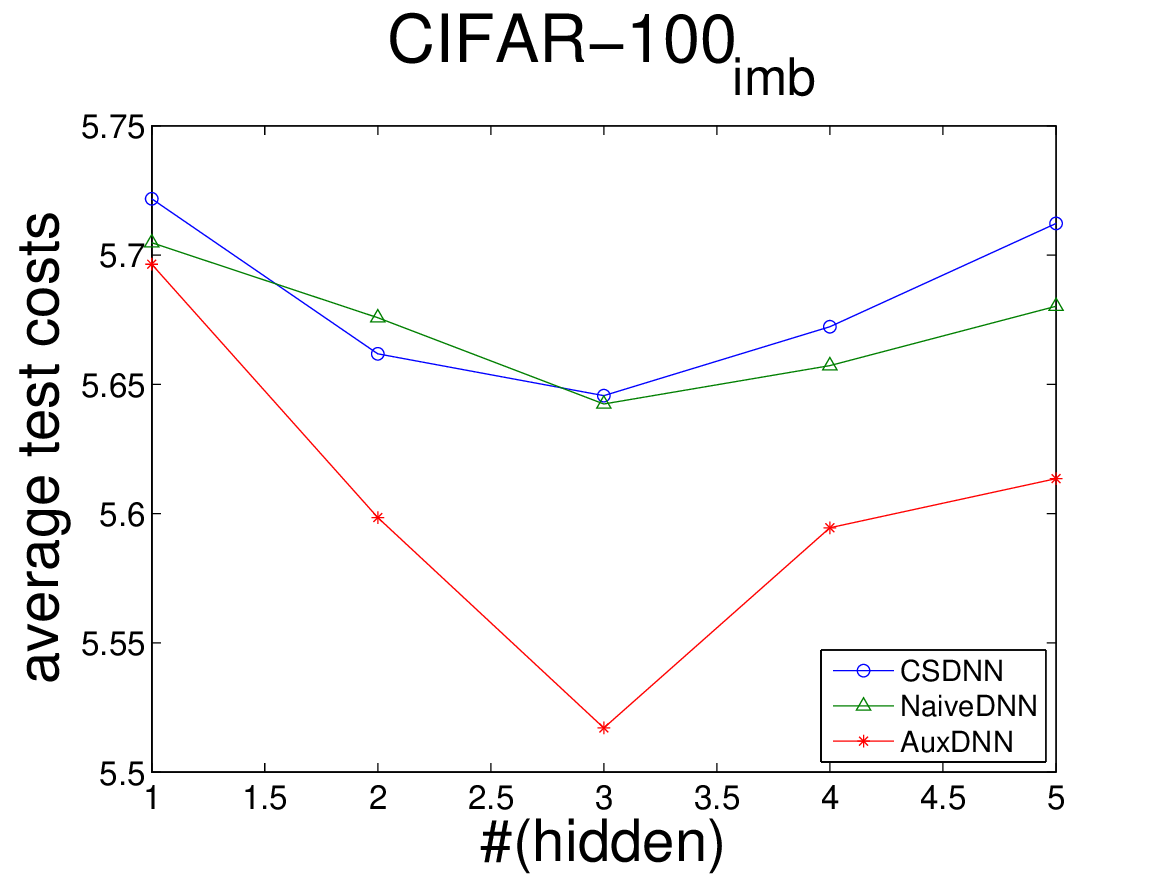}
    \end{minipage}\qquad
    \caption{
      This figure shows the results of our second experiment described in Section~\ref{sec:FC_compare}.
      The six sub-figures display the performance of the three competing DNNs on MNIST, CIFAR-10, CIFAR-100, $\mathrm{MNIST}_{\mathrm{imb}}$, $\mathrm{CIFAR}$-$\mathrm{10}_{\mathrm{imb}}$, and $\mathrm{CIFAR}$-$\mathrm{100}_{\mathrm{imb}}$, where each curve corresponds to one competitor.
      The x-axis is the number of hidden layers used to construct the DNN, and the y-axis is the corresponding average test costs.
    }
    \label{fig:hidden_plot}
  \end{figure*}

  The experimental results are displayed in Figure~\ref{fig:hidden_plot}.
  The x-axis is the number of hidden layers and the y-axis is the corresponding average test costs achieved.
  As we can observe from Figure~\ref{fig:hidden_plot}, when the number of hidden layers was less than or equal to three, CSDNN outperformed NaiveDNN probably because CSAE were doing cost-aware feature extraction relatively well, which accorded to the experimental results in~\cite{YC2016}.
  When the number of hidden layers exceeded three, all of the three models began to suffer from overfitting, causing their average test costs to increase.
  However, by looking at CSDNN and NaiveDNN, it was interesting to observe that although the average test costs of both models increased, the extent of increment of CSDNN was larger than that of NaiveDNN.
  We inferred that this phenomenon was ascribed to the diminishing gradients caused by sigmoid functions used in CSDNN, and although CSAE had done their best to mitigate this problem when the network was relatively shallow, CSDNN can still not escape the fate of diminishing gradients when the network grew deeper.
  This phenomenon could not be observed in~\cite{YC2016} because the deepest network they built had only three hidden layers.
  As for AuxDNN, it significantly outperformed both CSDNN and NaiveDNN regardless of the number of hidden layers, this further demonstrated the usefulness of our proposed AuxCST framework.

\begin{figure}[h]
    \centering
    \includegraphics[scale=0.36]{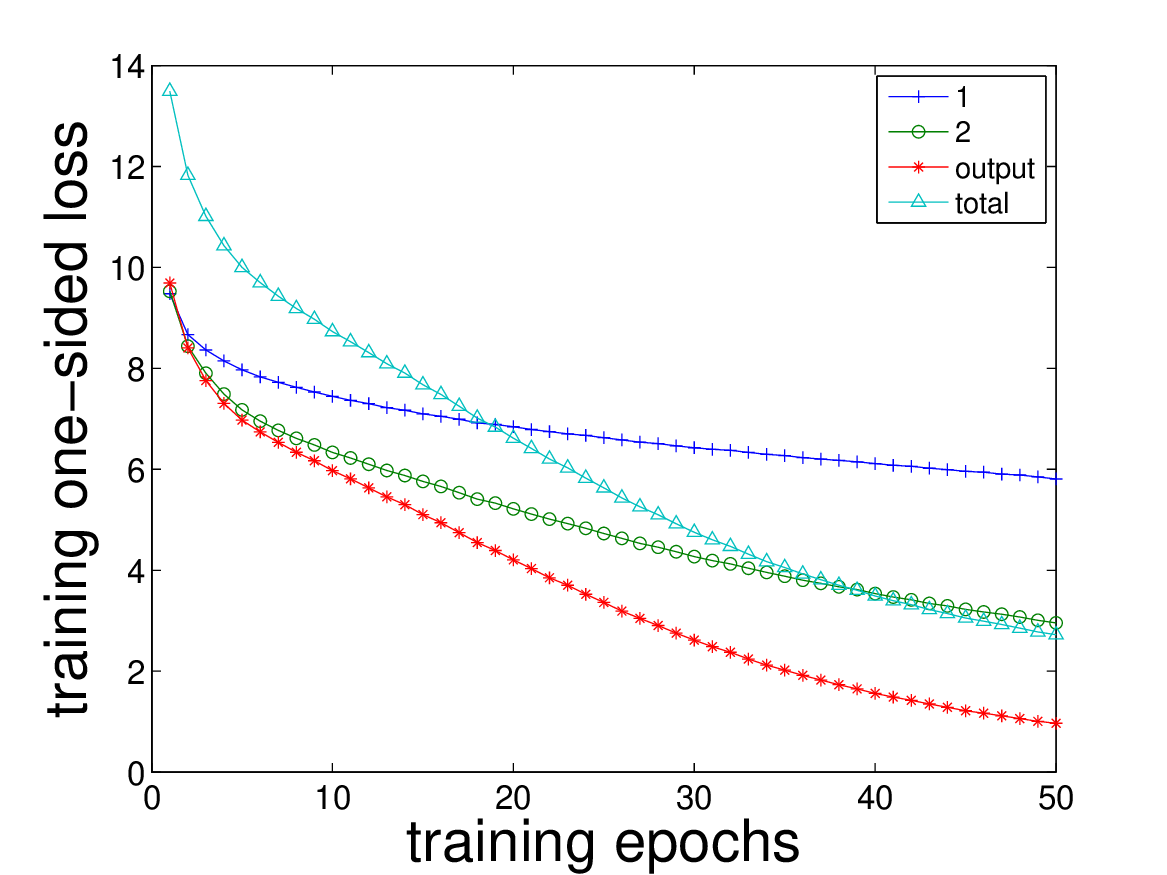}
    \caption{
      This figure shows the learning process of AuxDNN with three hidden layers on CIFAR-10.
      The blue and green represent the training loss of the first and second hidden layers, the red curve represents the training loss of of the output layer, and the aquamarine curve represents the total loss.
      The x-axis is the training epochs, and the y-axis is the corresponding one-sided loss.
    }
    \label{fig:learning_plot}
  \end{figure}

  To have a more comprehensive understanding on what the auxiliary outputs are doing, we also record the layer-wise one-sided training loss of AuxDNN with three hidden layers and plot the learning curve in Figure~\ref{fig:learning_plot} (we only plot the results of CIFAR-10, as the results of MNIST and CIFAR-100 are similar to CIFAR-10).
  Since the AuxDNN comes with three hidden layers, there would be two auxiliary outputs and one main output at the end.
  From the figure, we can observe that the earlier layers primarily serve as assistance, as their training loss decrease much slower than the last layer, which is the main target.

\subsection{Category Closeness as Cost Information}
  \label{sec:closeness}
  \begin{figure}
    \centering
    \includegraphics[scale=0.34]{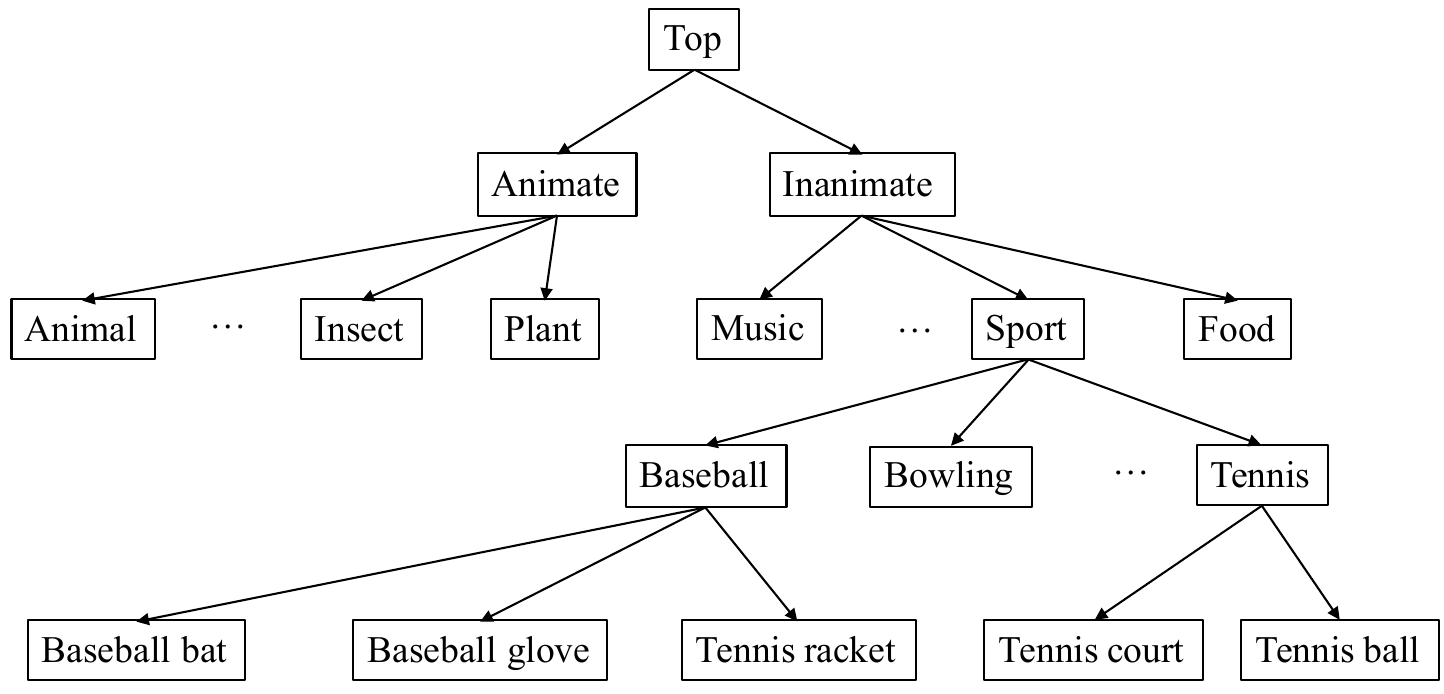}
    \caption{
      The figure shows a part of the hierarchy tree structure of the 256 classes in Caltech-256.
      The cost information is defined by the distance between classes.
      For example, the cost of classifying a baseball bat as a baseball glove is 2 since it takes only two steps to go from baseball bat to baseball glove according to the tree paths; while the cost of classifying a baseball bat as a tennis racket would be 4.
    }
    \label{fig:tree_example}
  \end{figure}

  In Section~\ref{sec:alpha_perf_relation} and~\ref{sec:FC_compare}, the cost information for all data sets are generated by randomized proportional setup.
  In our third experiment, instead of randomized proportional setup, we use a property featured in Caltech-256 data set to generate a much more interesting and realistic cost information.
  In Caltech-256, the 256 classes present a top-down hierarchical tree structure.
  Since the 256 classes present a hierarchy tree, the cost of classifying a class-A instance as class B could be interestingly and intuitively defined by the distance between class A and class B within the hierarchy tree.
  For example, in Figure~\ref{fig:tree_example}, the cost of classifying a baseball bat as a baseball glove is 2, since it takes only two steps to go from baseball bat to baseball glove through the tree paths; while the cost of classifying a baseball bat as a tennis racket is 4.

  \begin{figure}[]
    \centering
    \includegraphics[scale=0.32]{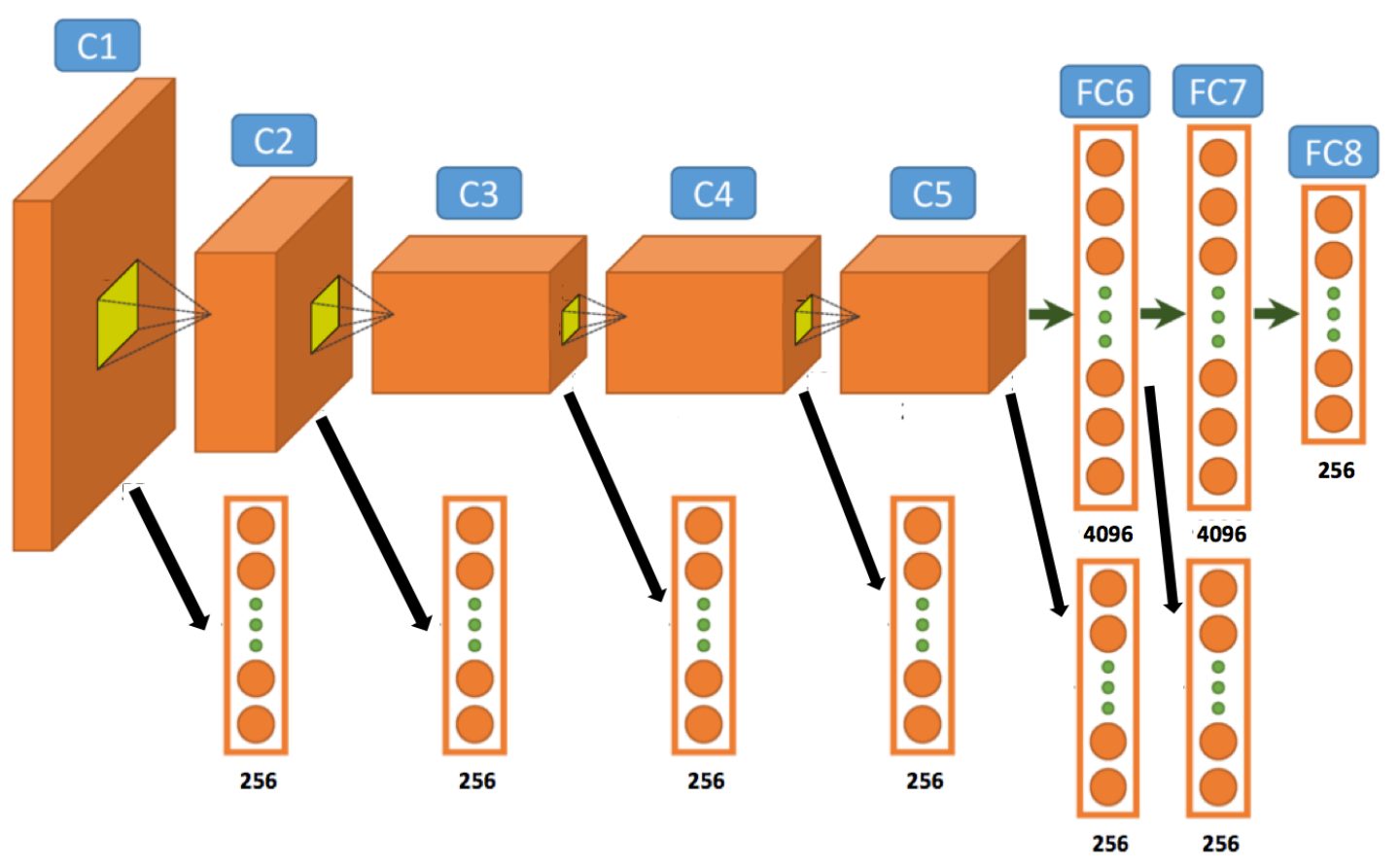}
    \caption{
      This figure shows how to apply the AuxCST framework to AlexNet.
      AlexNet consists of five convolutional layers and three fully-connected layers, where the last softmax layer is replaced with regression outputs for per-class cost estimation.
      For each hidden layer in AlexNet, 256 additional neurons are added as auxiliary per-class cost estimation.
      This network is referred to as AuxAlexNet in the experiment.
    }
    \label{fig:alexnet_auxit}
  \end{figure}

  We apply the AuxCST framework to the well-known AlexNet~\citep{krizhevsky2012imagenet} to tackle this cost-sensitive classification problem.
  AlexNet consists of five convolutional layers and three fully-connected layers.
  We use the off-the-shelf implementation of AlexNet provided in \cite{ding2014theano}, but replace the last softmax layer with regression outputs for per-class cost estimation so that we can apply $L_{\mathrm{OSR}}$ and the AuxCST framework.
  There are around 30K images in Caltech-256, we randomly split 20K images for training and the rest 10K images for testing.
  As there are 60 million parameters to be learned in AlexNet, a training set contains only 20K images turns out to be insufficient to learn so many parameters without considerable overfitting if we train the entire AlexNet on Caltech-256 from scratch.
  To deal with this problem, we first take AlexNet pretrained on ImageNet~\citep{ILSVRC15} provided in~\cite{ding2014theano} as initialization, then replace the softmax layer with regression outputs, and finally fine-tune the AlexNet on Caltech-256.
  The final AlexNet equipped with AuxCST framework is shown in Figure~\ref{fig:alexnet_auxit}, and will be referred to as AuxAlexNet.
  One may wonder that how come a DNN trained for a regular classification task can be transferred to a cost-sensitive classification one.
  However, this turns out to be not a problem, and will be discussed later.

  \begin{figure}[h]
    \centering
    \includegraphics[scale=0.32]{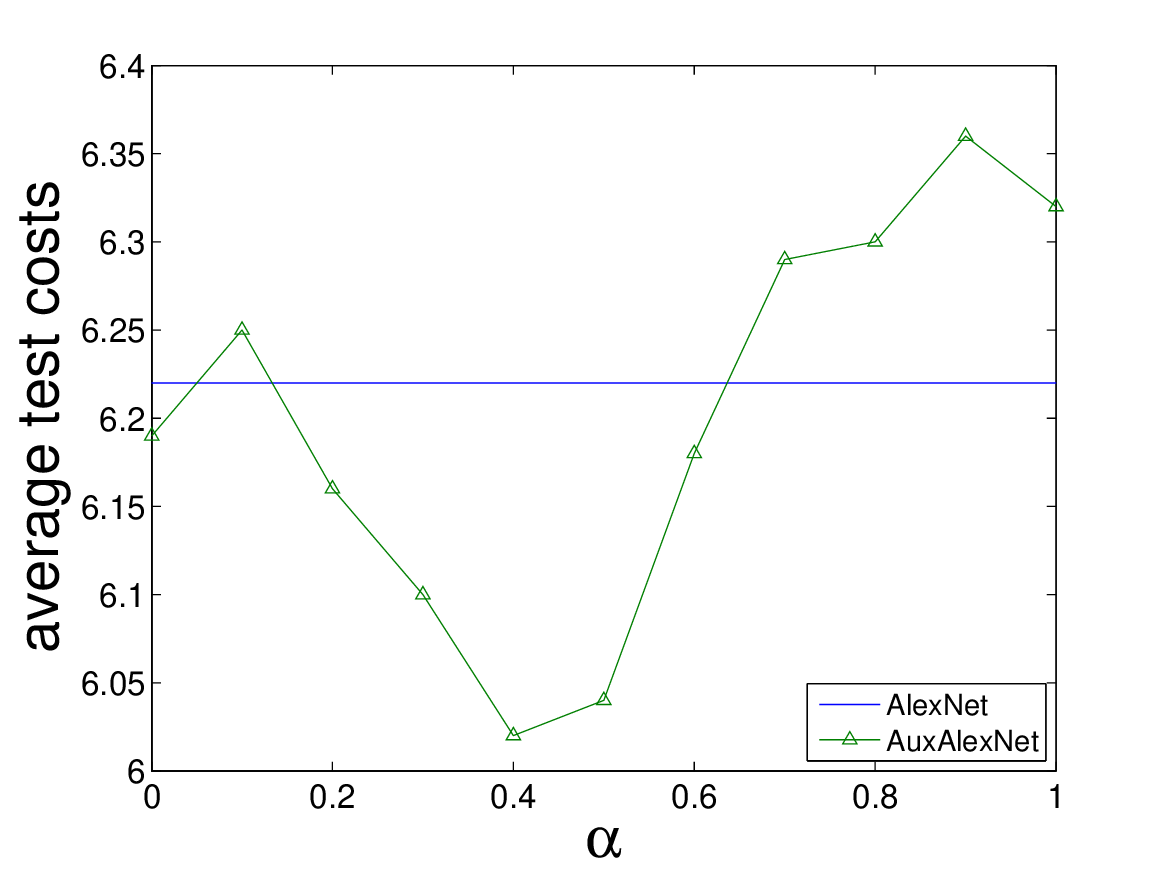}
    \caption{
      This figure shows the results of our third experiment described in Section~\ref{sec:closeness}.
      The x-axis is the value of $\alpha$ in~(\ref{eq:same_alpha}) and is specific to AuxAlexNet, which is represented by the green curve, and the y-axis is the corresponding average test costs.
      The blue curve represents the cost-blind AlexNet and has nothing to do with~$\alpha$.
    }
    \label{fig:closeness_results}
  \end{figure}

  We compare AuxAlexNet with the original AlexNet, which is set as the cost-blind deep learning baseline.
  The original AlexNet is also first pretrained on ImageNet, then fine-tuned on Caltech-256.
  Similar to the first experiment in Section~\ref{sec:alpha_perf_relation}, we will also try varying values for $\alpha$ to observe how $\alpha$ interacts with the performance of AuxAlexNet.
  The experimental results are shown in Figure~\ref{fig:closeness_results}.
  As we can see from the results, AuxAlexNet achieved very different average test costs with different $\alpha$, but still outperformed the original AlexNet in most cases.
  It is actually not surprising to see that the transfer learning from a regular classification task to a cost-sensitive classification one works quite well.
  Recall that the function of DNN is feature extraction, and the cost information in a cost-sensitive classification task just further asks the DNN to pay more attention on some specific features to prevent misclassifying those examples that belong to some really important classes (will need to pay relatively high cost if misclassify them).
  In this experiment, we give an example of how to apply the proposed AuxCST framework to a modern network to tackle cost-sensitive classification problem, and the experimental results on Caltech-256 with category closeness as cost information validate the usefulness of the AuxCST framework.

\section{Conclusion and Future Work}
  \label{sec:con}
  We propose a novel framework Auxiliary Cost-Sensitive Targets~(AuxCST) for general end-to-end cost-sensitive deep learning.
  Different from the previous approaches, the framework can be applied to DNN that consists of any structures to tackle challenging cost-sensitive classification problems.
  Extensive experimental results on MNIST, CIFAR-10, CIFAR-100, and Caltech-256 with two cost information settings demonstrate the usefulness of the proposed framework for making any advanced DNN models cost-sensitive.
  In the future, we will build a deeper network with AuxCST framework to tackle ImageNet cost-sensitive classification problem.


\bibliographystyle{IEEEtranN}
\bibliography{mybib}

\begin{thebibliography}{39}
\providecommand{\natexlab}[1]{#1}
\providecommand{\url}[1]{#1}
\csname url@samestyle\endcsname
\providecommand{\newblock}{\relax}
\providecommand{\bibinfo}[2]{#2}
\providecommand{\BIBentrySTDinterwordspacing}{\spaceskip=0pt\relax}
\providecommand{\BIBentryALTinterwordstretchfactor}{4}
\providecommand{\BIBentryALTinterwordspacing}{\spaceskip=\fontdimen2\font plus
\BIBentryALTinterwordstretchfactor\fontdimen3\font minus
  \fontdimen4\font\relax}
\providecommand{\BIBforeignlanguage}[2]{{%
\expandafter\ifx\csname l@#1\endcsname\relax
\typeout{** WARNING: IEEEtranN.bst: No hyphenation pattern has been}%
\typeout{** loaded for the language `#1'. Using the pattern for}%
\typeout{** the default language instead.}%
\else
\language=\csname l@#1\endcsname
\fi
#2}}
\providecommand{\BIBdecl}{\relax}
\BIBdecl

\bibitem[Krizhevsky et~al.(2012)Krizhevsky, Sutskever, and
  Hinton]{krizhevsky2012imagenet}
A.~Krizhevsky, I.~Sutskever, and G.~E. Hinton, ``Imagenet classification with
  deep convolutional neural networks,'' in \emph{NeurIPS}, 2012.

\bibitem[Ciregan et~al.(2012)Ciregan, Meier, and Schmidhuber]{ciregan2012multi}
D.~Ciregan, U.~Meier, and J.~Schmidhuber, ``Multi-column deep neural networks
  for image classification,'' in \emph{CVPR}, 2012.

\bibitem[Simonyan and Zisserman(2015)]{simonyan2014very}
K.~Simonyan and A.~Zisserman, ``Very deep convolutional networks for
  large-scale image recognition,'' 2015.

\bibitem[Szegedy et~al.(2015)Szegedy, Liu, Jia, Sermanet, Reed, Anguelov,
  Erhan, Vanhoucke, and Rabinovich]{szegedy2015going}
C.~Szegedy, W.~Liu, Y.~Jia, P.~Sermanet, S.~Reed, D.~Anguelov, D.~Erhan,
  V.~Vanhoucke, and A.~Rabinovich, ``Going deeper with convolutions,'' in
  \emph{CVPR}, 2015.

\bibitem[He et~al.(2016)He, Zhang, Ren, and Sun]{he2016deep}
K.~He, X.~Zhang, S.~Ren, and J.~Sun, ``Deep residual learning for image
  recognition,'' in \emph{CVPR}, 2016.

\bibitem[Hinton et~al.(2012)Hinton, Deng, Yu, Dahl, Mohamed, Jaitly, Senior,
  Vanhoucke, Nguyen, Sainath, et~al.]{hinton2012deep}
G.~Hinton, L.~Deng, D.~Yu, G.~Dahl, A.-r. Mohamed, N.~Jaitly, A.~Senior,
  V.~Vanhoucke, P.~Nguyen, T.~Sainath \emph{et~al.}, ``Deep neural networks for
  acoustic modeling in speech recognition: The shared views of four research
  groups,'' \emph{IEEE Signal Processing Magazine}, vol.~29, no.~6, pp. 82--97,
  2012.

\bibitem[Dahl et~al.(2012)Dahl, Yu, Deng, and Acero]{dahl2012context}
G.~Dahl, D.~Yu, L.~Deng, and A.~Acero, ``Context-dependent pre-trained deep
  neural networks for large-vocabulary speech recognition,'' \emph{TASLP},
  vol.~20, no.~1, pp. 30--42, 2012.

\bibitem[Tan(1993)]{tan1993cost}
M.~Tan, ``Cost-sensitive learning of classification knowledge and its
  applications in robotics,'' \emph{Machine Learning}, vol.~13, no.~1, pp.
  7--33, 1993.

\bibitem[Chan and Stolfo(1998)]{chan1998toward}
P.~K. Chan and S.~J. Stolfo, ``Toward scalable learning with non-uniform class
  and cost distributions: {A} case study in credit card fraud detection,'' in
  \emph{KDD}, 1998.

\bibitem[Fan et~al.(2000)Fan, Lee, Stolfo, and Miller]{fan2000multiple}
W.~Fan, W.~Lee, S.~Stolfo, and M.~Miller, ``A multiple model cost-sensitive
  approach for intrusion detection,'' in \emph{ECML}, 2000.

\bibitem[Zhang and Zhou(2010)]{zhang2010cost}
Y.~Zhang and Z.-H. Zhou, ``Cost-sensitive face recognition,'' \emph{TPAMI},
  vol.~32, no.~10, pp. 1758--1769, 2010.

\bibitem[Jan et~al.(2011)Jan, Lin, Chen, Chern, Huang, Wen, Chung, Li, Chuang,
  Li, Chan, Wang, Wang, Lin, and Wang]{TJ2011}
T.-K. Jan, H.-T. Lin, H.-P. Chen, T.-C. Chern, C.-Y. Huang, B.-C. Wen, C.-W.
  Chung, Y.-J. Li, Y.-C. Chuang, L.-L. Li, Y.-J. Chan, J.-K. Wang, Y.-L. Wang,
  C.-H. Lin, and D.-W. Wang, ``Cost-sensitive classification on pathogen
  species of bacterial meningitis by {Surface Enhanced Raman Scattering},'' in
  \emph{BIBM}, 2011.

\bibitem[Lu and Tan(2010)]{lu2010cost}
J.~Lu and Y.-P. Tan, ``Cost-sensitive subspace learning for face recognition,''
  in \emph{CVPR}, 2010.

\bibitem[Zhang et~al.(2014)Zhang, Li, Zhou, Huang, and Shang]{zhang2014cost}
L.~Zhang, H.~Li, X.~Zhou, B.~Huang, and L.~Shang, ``Cost-sensitive sequential
  three-way decision for face recognition,'' in \emph{RSEISP}, 2014.

\bibitem[Zhang et~al.(2016)Zhang, Sun, Ji, Yuan, and Sun]{zhang2016cost}
G.~Zhang, H.~Sun, Z.~Ji, Y.-H. Yuan, and Q.~Sun, ``Cost-sensitive dictionary
  learning for face recognition,'' \emph{Pattern Recognition}, vol.~60, pp.
  613--629, 2016.

\bibitem[Kukar and Kononenko(1998)]{kukar1998cost}
M.~Kukar and I.~Kononenko, ``Cost-sensitive learning with neural networks,'' in
  \emph{ECAI}, 1998.

\bibitem[Domingos(1999)]{domingos1999metacost}
P.~Domingos, ``Metacost: A general method for making classifiers
  cost-sensitive,'' in \emph{KDD}, 1999.

\bibitem[Zadrozny and Elkan(2001)]{zadrozny2001learning}
B.~Zadrozny and C.~Elkan, ``Learning and making decisions when costs and
  probabilities are both unknown,'' in \emph{KDD}, 2001.

\bibitem[Tu and Lin(2010)]{HT2010}
H.-H. Tu and H.-T. Lin, ``One-sided support vector regression for multi-class
  cost-sensitive classification,'' in \emph{ICML}, 2010.

\bibitem[Zhou and Liu(2006)]{zhou2006training}
Z.-H. Zhou and X.-Y. Liu, ``Training cost-sensitive neural networks with
  methods addressing the class imbalance problem,'' \emph{TKDE}, vol.~18,
  no.~1, pp. 63--77, 2006.

\bibitem[Chung et~al.(2016)Chung, Lin, and Yang]{YC2016}
Y.-A. Chung, H.-T. Lin, and S.-W. Yang, ``Cost-aware pre-training for
  multiclass cost-sensitive deep learning,'' in \emph{IJCAI}, 2016.

\bibitem[Bengio(2009)]{bengio2009learning}
Y.~Bengio, ``Learning deep architectures for ai,'' \emph{Machine Learning},
  vol.~2, no.~1, pp. 1--127, 2009.

\bibitem[LeCun et~al.(1998)LeCun, Bottou, Bengio, and
  Haffner]{lecun1998gradient}
Y.~LeCun, L.~Bottou, Y.~Bengio, and P.~Haffner, ``Gradient-based learning
  applied to document recognition,'' \emph{Proceedings of the IEEE}, vol.~86,
  no.~11, pp. 2278--2324, 1998.

\bibitem[Nair and Hinton(2010)]{nair2010rectified}
V.~Nair and G.~Hinton, ``Rectified linear units improve restricted boltzmann
  machines,'' in \emph{ICML}, 2010.

\bibitem[Lee et~al.(2015)Lee, Xie, Gallagher, Zhang, and Tu]{lee2015deeply}
C.-Y. Lee, S.~Xie, P.~Gallagher, Z.~Zhang, and Z.~Tu, ``Deeply-supervised
  nets,'' in \emph{AISTATS}, 2015.

\bibitem[Teerapittayanon et~al.(2016)Teerapittayanon, McDanel, and
  Kung]{teerapittayanon2016branchynet}
S.~Teerapittayanon, B.~McDanel, and H.~Kung, ``Branchynet: Fast inference via
  early exiting from deep neural networks,'' in \emph{ICPR}, 2016.

\bibitem[Abe et~al.(2004)Abe, Zadrozny, and Langford]{abe2004iterative}
N.~Abe, B.~Zadrozny, and J.~Langford, ``An iterative method for multi-class
  cost-sensitive learning,'' in \emph{KDD}, 2004.

\bibitem[Dayhoff(1990)]{dayhoff1990intro}
J.~Dayhoff, \emph{Neural Network Architectures: An Introduction}.\hskip 1em
  plus 0.5em minus 0.4em\relax Van Nostrand Reinhold Co., 1990.

\bibitem[Lawrence et~al.(1997)Lawrence, Giles, Tsoi, and
  Back]{lawrence1997face}
S.~Lawrence, L.~Giles, A.~C. Tsoi, and A.~Back, ``Face recognition: A
  convolutional neural-network approach,'' \emph{TNN}, vol.~8, no.~1, pp.
  98--113, 1997.

\bibitem[Hinton et~al.(2006)Hinton, Osindero, and Teh]{hinton2006fast}
G.~Hinton, S.~Osindero, and Y.-W. Teh, ``A fast learning algorithm for deep
  belief nets,'' \emph{Neural Computation}, vol.~18, no.~7, pp. 1527--1554,
  2006.

\bibitem[Nielsen(2015)]{michael2015nndl}
M.~Nielsen, \emph{Neural Networks and Deep Learning}.\hskip 1em plus 0.5em
  minus 0.4em\relax Determination Press, 2015.

\bibitem[Glorot et~al.(2011)Glorot, Bordes, and Bengio]{glorot2011deep}
X.~Glorot, A.~Bordes, and Y.~Bengio, ``Deep sparse rectifier neural networks,''
  in \emph{AISTATS}, 2011.

\bibitem[Xu et~al.(2015)Xu, Wang, Chen, and Li]{xu2015empirical}
B.~Xu, N.~Wang, T.~Chen, and M.~Li, ``Empirical evaluation of rectified
  activations in convolutional network,'' \emph{arXiv preprint
  arXiv:1505.00853}, 2015.

\bibitem[He et~al.(2015)He, Zhang, Ren, and Sun]{he2015delving}
K.~He, X.~Zhang, S.~Ren, and J.~Sun, ``Delving deep into rectifiers: Surpassing
  human-level performance on imagenet classification,'' in \emph{ICCV}, 2015.

\bibitem[Maas et~al.(2013)Maas, Hannun, and Ng]{maas2013rectifier}
A.~L. Maas, A.~Y. Hannun, and A.~Y. Ng, ``Rectifier nonlinearities improve
  neural network acoustic models,'' in \emph{ICML, Workshop on Deep Learning
  for Audio, Speech and Language Processing}, 2013.

\bibitem[Krizhevsky and Hinton(2009)]{Krizhevsky09}
A.~Krizhevsky and G.~Hinton, ``Learning multiple layers of features from tiny
  images,'' \emph{Master's thesis, Department of Computer Science, University
  of Toronto}, 2009.

\bibitem[Griffin et~al.(2007)Griffin, Holub, and Perona]{griffin2007caltech}
G.~Griffin, A.~Holub, and P.~Perona, ``Caltech-256 object category dataset,''
  \emph{California Institute of Technology}, 2007.

\bibitem[Ding et~al.(2015)Ding, Wang, Mao, and Taylor]{ding2014theano}
W.~Ding, R.~Wang, F.~Mao, and G.~Taylor, ``Theano-based large-scale visual
  recognition with multiple gpus,'' in \emph{ICLR}, 2015.

\bibitem[Russakovsky et~al.(2015)Russakovsky, Deng, Su, Krause, Satheesh, Ma,
  Huang, Karpathy, Khosla, Bernstein, Berg, and Fei-Fei]{ILSVRC15}
O.~Russakovsky, J.~Deng, H.~Su, J.~Krause, S.~Satheesh, S.~Ma, Z.~Huang,
  A.~Karpathy, A.~Khosla, M.~Bernstein, A.~C. Berg, and L.~Fei-Fei, ``Imagenet
  large scale visual recognition challenge,'' \emph{International Journal of
  Computer Vision}, vol. 115, no.~3, pp. 211--252, 2015.

\end{thebibliography}

\end{document}